\documentclass[11pt]{article}

\usepackage{microtype}
\usepackage{graphicx}
\usepackage{subcaption}
\usepackage{booktabs} 

\usepackage{hyperref}
\usepackage{fontawesome5}

\usepackage[final]{acl}
\usepackage{times}
\usepackage{latexsym}

\usepackage[T1]{fontenc}

\usepackage[utf8]{inputenc}

\usepackage{microtype}

\usepackage{inconsolata}
\usepackage{amssymb}
\usepackage{mathtools}
\usepackage{amsthm}
\usepackage{thmtools}
\usepackage{thm-restate}
\usepackage{chngcntr}
\usepackage{booktabs}
\usepackage{multirow}
\usepackage[table, xcdraw]{xcolor} 
\usepackage{graphicx} 
\usepackage{placeins}
\usepackage{listings}
\usepackage[capitalize,noabbrev]{cleveref}

\theoremstyle{plain}
\newtheorem{theorem}{Theorem}[section]
\newtheorem{proposition}[theorem]{Proposition}

\theoremstyle{definition}
\newtheorem{definition}[theorem]{Definition}

\theoremstyle{remark}

\usepackage[disable,textsize=tiny]{todonotes}

\newcommand{\blambda}{\boldsymbol{\lambda}}

\newcommand{\bbR}{\mathbb{R}}

\newcommand{\cM}{\mathcal{M}}

\newcommand{\cA}{\mathcal{A}}

\makeatother

\counterwithout{proposition}{section}
\counterwithout{lemma}{section}
\counterwithout{theorem}{section}

\newcommand{\think}[1]{\textcolor{blue}{\texttt{<think>}} #1 \textcolor{blue}{\texttt{</think>}}}
\newcommand{\search}[1]{\textcolor{cyan}{\texttt{<search>}} #1 \textcolor{cyan}{\texttt{</search>}}}
\newcommand{\info}[1]{\textcolor{brown}{\texttt{<information>}} #1 \textcolor{brown}{\texttt{</information>}}}
\newcommand{\answer}[1]{\textcolor{purple}{\texttt{<answer>}} #1 \textcolor{purple}{\texttt{</answer>}}}
\newcommand{\sql}[1]{\textcolor{red}{\texttt{<sql>}} #1 \textcolor{red}{\texttt{</sql>}}}
\newcommand{\solution}[1]{\textcolor{green}{\texttt{<solution>}} #1 \textcolor{green}{\texttt{</solution>}}}

\title{Reasoning and Tool-use Compete in Agentic RL:\\From Quantifying Interference to Disentangled Tuning}

\author{%
Yu Li$^{1}$, Mingyang Yi$^{1}$\thanks{Corresponding author.}, Xiuyu Li$^{1}$, Ju Fan$^{1}$, \\
\textbf{Fuxin Jiang$^{2}$, Binbin Chen$^{2}$, Peng Li$^{2}$, Jie Song$^{2}$, Tieying Zhang$^{2}$\footnotemark[1]} \\
$^{1}$School of Information, Renmin University of China 
$^{2}$Bytedance Inc.\\
\texttt{yimingyang@ruc.edu.cn}, \quad \texttt{tieying.zhang@bytedance.com}
}

\begin{document}
\maketitle
\begin{abstract}
Agentic Reinforcement Learning (ARL) trains large language models to interleave reasoning with external tool execution to solve complex tasks.
Most existing ARL methods train a single set of parameters to support both reasoning and tool-use behaviors, implicitly assuming that joint training leads to improved overall agent performance. 
Despite its widespread adoption, this assumption has rarely been examined empirically.
In this paper, we systematically examine this assumption by introducing Capability Effect Attribution (CEA), which provides quantitative evidence of interference between reasoning and tool-use behaviors.
Through an in-depth analysis, we show that these two capabilities often induce misaligned gradient directions, leading to training interference that undermines the effectiveness of joint optimization and challenges the prevailing ARL paradigm. 
To address this issue, we propose Disentangled Action–Reasoning Tuning (DART), a simple and efficient framework that explicitly decouples parameter updates for reasoning and tool use via separate low-rank adaptation modules.
With this simple change alone, DART outperforms all joint-optimization baselines and approaches the 2-Agent upper bound across thirteen benchmarks on retrieval-augmented QA and NL2SQL, further supporting our finding of capability interference under shared optimization.\footnote{Code is available at \href{https://github.com/liyu199809/Disentangled-Action-Reasoning-Tuning-for-Agentic-RL}{\faGithub~\texttt{liyu199809/DART}}.}
\end{abstract}

\section{Introduction}
\label{introduction}
Recent advances in Agentic Reinforcement Learning (ARL) for post-training~\cite{ouyang2022training,bai2022constitutional,li2026ets,liu2026automated} have substantially extended the capabilities of large language models (LLMs).
Beyond text generation, modern LLMs can perform complex reasoning and interact with external tools~\cite{liu2026toolanchor} to solve tasks such as information retrieval~\cite{jin2025searchr}, computation~\cite{mai2025agentic}, data analysis~\cite{zhang2025deepanalyze}, and research workflows~\cite{qiao2025webresearcher}.

The goal of ARL is to train models that reliably execute external tools while exhibiting strong reasoning abilities~\cite{wu2024avatar}.
Most existing ARL paradigms~\cite{chai2025rlfactory,shao2024deepseekmath,zeng2024agenttuning,zhang2025reward} jointly optimize these two \emph{heterogeneous capabilities} based on a \emph{single ARL objective with shared model parameters.} 
This design implicitly assumes that tool execution and logical reasoning can be effectively accommodated within the same parameter space.
Prior work has shown that optimization interference arises when training across distinct domains~\cite{zhao2025redone,yuan2026differential,wu2025masksearch}; however, whether heterogeneous capabilities \emph{within} a single agentic domain also interfere with each other remains largely unexplored.

In this work, we directly test the shared parameter assumption through a controlled empirical analysis of the interaction between tool-use and reasoning.
Specifically, we introduce \textbf{Capability Effect Attribution} (CEA), a diagnostic framework that decomposes an agent's performance into individual capability effects and pairwise interaction terms.
By constructing six controlled model variants via gradient masking and hybrid inference, we solve for per-question interaction coefficients and reveal a significant \textbf{negative interaction} between reasoning and tool-use under joint optimization, indicating that shared-parameter training induces implicit competition.

To explain the root cause of this interference, we examine the optimization dynamics~\citep{ren2025learning,li2026towards} by analyzing gradients from reasoning and tool-use tokens. We identify a clear gradient misalignment: the two types of gradients are nearly orthogonal, causing joint optimization over a shared backbone to update parameters in a \textbf{compromise direction} that is suboptimal for both capabilities.

Motivated by this finding, we propose \textbf{Disentangled Action-Reasoning Tuning (DART)}, a simple yet effective framework designed to test whether eliminating gradient misalignment improves performance.
DART freezes the pretrained backbone and routes reasoning and tool-use tokens to separate LoRA~\cite{hulora} adapters, so that each capability updates only its own parameters without affecting the other.

Experiments on seven QA and six NL2SQL benchmarks show that DART consistently outperforms joint-optimization baselines, confirming the generality of capability interference and the effectiveness of gradient disentanglement.
Two experimental observations further support this conclusion:
\textbf{(1)} DART surpasses all joint-optimization baselines, including multi-LoRA methods, showing that resolving gradient misalignment via hard token-level routing is essential.
\textbf{(2)} Ablation studies attribute DART's gains to misalignment elimination rather than extra parameter capacity.

Our contributions are summarized as follows:
\begin{itemize}
    \vspace{-2mm}
    \item For ARL training, we empirically identify negative interaction between tool-use and reasoning using Capability Effect Attribution (CEA), and trace this interference to gradient misalignments under joint optimization.
    \vspace{-2mm}
    \item To validate this finding, we propose DART, a simple yet effective framework that disentangles gradients for reasoning and tool-use via separate LoRA adapters.
    \vspace{-2mm}
    \item Extensive experiments on two tasks across thirteen benchmarks show that DART surpasses joint-optimization baselines, confirming that capability interference is a non-negligible bottleneck in ARL and that gradient disentanglement is an effective remedy.
\end{itemize}

\section{Related Work}
\paragraph{ARL with Tool-use.}
ARL research focuses on fine-tuning LLMs as autonomous agents that learn to invoke external tools through environment feedback, bridging the gap between reasoning and action without dense step-level supervision tuning~\cite{schick2023toolformer}.
Recent advancements have optimized various components of this pipeline, including reward formulation to induce emergent behaviors~\cite{qian2025toolrl,peiyuan2024agile,mai2025agentic}, policy refinement for precise action interleaving~\cite{feng2025retool,singh2025agentic,wei2025autotir} , and large-scale trajectory synthesis~\cite{dong2025tool, li2025websailor} for scalable training~\cite{jiang2025verltool}. 
However, none of these works examine whether reasoning and tool-use interfere with each other under joint optimization, which is the central question of our study.

\paragraph{Multi-LoRA.}
Existing Multi-LoRA methods follow the MoE paradigm~\citep{shazeer2017outrageously}, either using a soft router to mix multiple adapters for greater capacity~\cite{li2024mixlora,luo2024moelora,zhu2023sira,wumixture,luo2025tr}, or composing adapters to generalize across domains~\cite{huanglorahub, wang2024customizable, ma2024modula}. 
These methods address a fundamentally different problem from ours: they aim to increase capacity or enable multi-domain transfer, not to prevent optimization interference between heterogeneous capabilities in a single domain setting.
Moreover, their \emph{soft mixing} routes each token's gradient to \emph{all} adapters, resulting in an interaction between them. 
In contrast, DART uses a deterministic hard router so each token updates exactly one adapter, preventing reasoning and tool-use from competing over shared trainable parameters.


\vspace{-1mm}
\section{Preliminaries}
\label{sec:preliminaries}


This section presents the Agentic Reinforcement Learning  and describes low-rank adaptation.

\subsection{Agentic Reinforcement Learning (ARL)}
An LLM agent $\pi_\theta(c_t \mid c_{<t})$ generates a trajectory $\tau$ under query $q$, interleaving reasoning and tool-use tokens. 
\begin{equation}
    \tau = (c_1, \dots, c_t, \dots, c_T),
\end{equation}
To distinguish the roles of tokens within a trajectory, we define a \textit{role-based} router function $\ell: \{1, \dots, T\} \to \{r, a\}$.
Here, $\ell(t) = r$ indicates that $c_t$ is a \textbf{reasoning token}, while $\ell(t) = a$ indicates a \textbf{tool-use token}.
Concretely, the router is defined as:
\begin{equation}
\ell(t)=
\begin{cases}
a, & \text{if } c_{<t} \text{ in tool-call span},\\
r, & \text{otherwise}.
\end{cases}
\label{eq:token_router}
\end{equation}
This assignment is fully deterministic and triggered by special tokens (e.g., \texttt{<search>} marks the start of a tool-call span as illustrated in Fig.~\ref{fig:grad+router}(B)).
The agent is optimized to maximize the expected reward $\mathcal{J}(\theta) = \mathbb{E}[R(\tau)]$.
We estimate the policy gradient:
\begin{small}
   \begin{equation}
    \nabla_\theta \mathcal{J}(\theta) \approx \mathbb{E}_{\tau \sim \pi_\theta} \left[ \mathcal{A}(\tau) \sum_{t=1}^T \nabla_\theta \log \pi_\theta(c_t \mid c_{<t}) \right],
    \label{eq:joint_gradient}
\end{equation} 
\end{small}
where $\mathcal{A}(\tau)$ is the advantage derived by reward $R(\tau)$.
In standard ARL, a single set of shared parameters $\theta$ is updated using gradients from both reasoning and tool-use tokens, without considering the distinction specified by $\ell(t)$.

\vspace{-2mm}
\subsection{Low-Rank Adaptation (LoRA)}
To reduce fine-tuning overhead, Low-Rank Adaptation (LoRA) freezes the pre-trained weights $W \in \mathbb{R}^{d \times h}$ and introduces trainable low-rank decomposition matrices.
For a given layer, let $\mathbf{h}_t \in \mathbb{R}^h$ denote the hidden state corresponding to token $c_t$. The forward pass is modified as:
\begin{equation}
    \mathbf{h}'_t = W \mathbf{h}_t + \Delta W \mathbf{h}_t = W \mathbf{h}_t + BA \mathbf{h}_t,
\end{equation}
where $B \in \mathbb{R}^{d \times r}$ and $A \in \mathbb{R}^{r \times h}$ are low-rank matrices with $r \ll \min(d, h)$.
Only $A$ and $B$ are trained while $W$ is frozen, and the same $\Delta W$ is applied to all tokens in $\tau$.

\section{Do Reasoning and Tool-Use Conflict?}
\label{sec:diagnosis}

\begin{figure*}[t]
    \centering
    \includegraphics[width=0.9\textwidth]{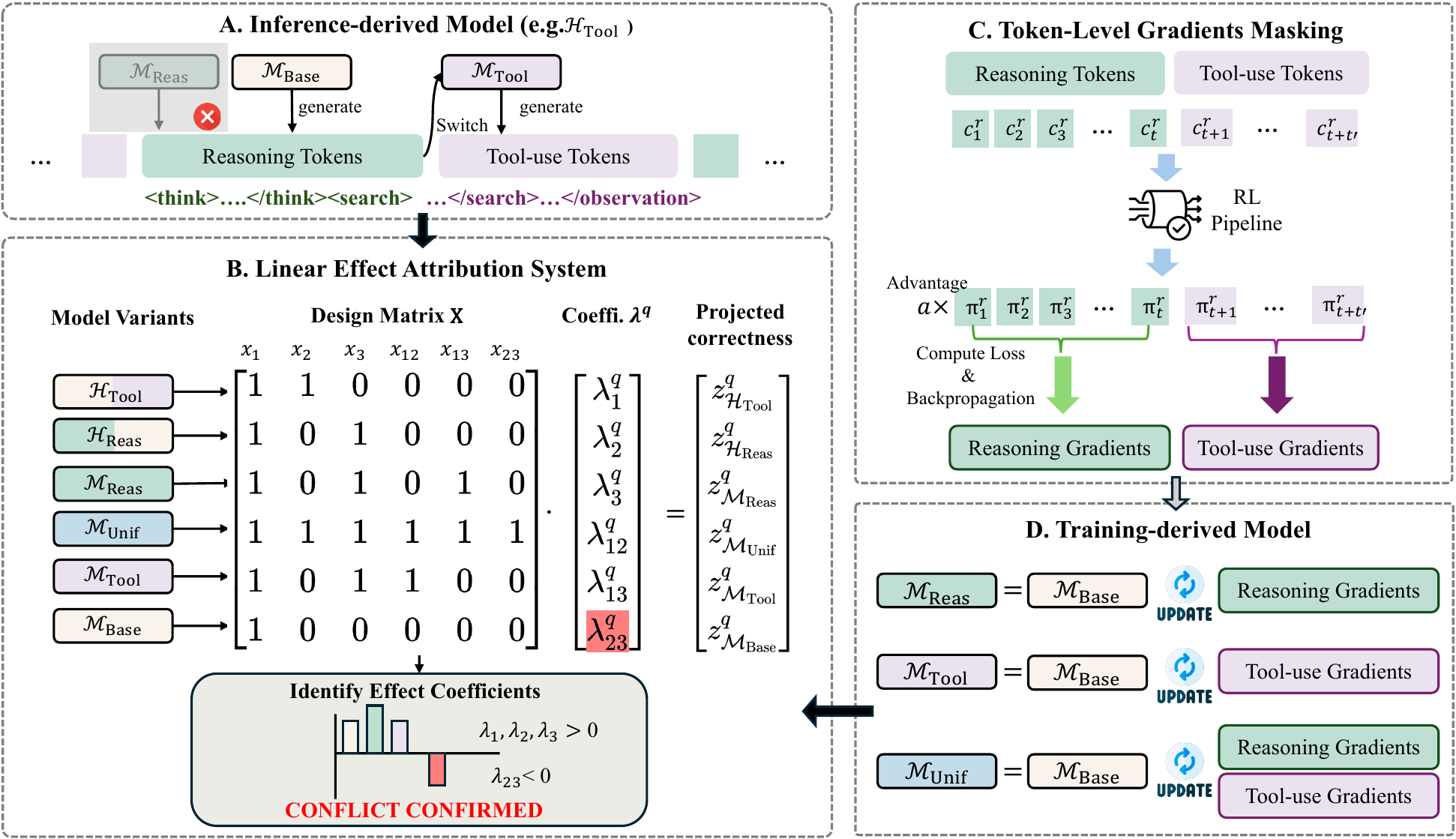}
    \caption{
    Overview of Capability Effect Attribution (CEA).
    \textbf{(A).Inference-derived Model:} Different token types are routed to separately trained models at inference time, composing capabilities without parameter-level interaction.
    \textbf{(B).Design Matrix and Attribution:} Six model variants populate the design matrix $\mathbf{X}$; solving the system yields per-question coefficients $\boldsymbol{\lambda}^q$, where $\lambda_{23}<0$ signals interference.
    \textbf{(C).Token-Level Gradient Masking:} Binary masks gate per-token gradient contributions, isolating capability-specific parameter updates during training.
    \textbf{(D).Training-derived Models:} Gradient masking produces specialized variants from a shared backbone, enabling controlled comparisons across capability configurations.
    }
    \vspace{-4mm}
    \label{fig:section3_main}
\end{figure*}

In this section, we investigate whether jointly optimizing reasoning and tool-use in ARL leads to interference.
We first introduce \emph{Capability Effect Attribution} (CEA), a diagnostic framework that selectively activates each capability via gradient masking and hybrid inference (\S~\ref{sec:capability_decomposition}--\ref{sec:gradient_masking}).
CEA reveals a clear negative interaction between reasoning and tool-use (\S~\ref{sec:empirical_analysis}), which we trace to gradient misalignment between the two token types (\S~\ref{sec:grad_conflict}).

\subsection{Formalizing Capability Effect Attribution}
\label{sec:capability_decomposition}

We formalize CEA by attributing a model's expected correctness to individual capability effects and pairwise interaction terms.
\begin{definition}
\label{def:capabilities}
The capabilities of agent are indicated by three binary indicators: base $x_1$, tool-use $x_2$, reasoning $x_3$. Here $x_{i} = 1$ indicates the capability exists and $x_{i} = 0$ otherwise. 
\end{definition}
\begin{definition}
Pairwise interaction indicators $x_{ij}$ ($i, j\in \{1, 2, 3\}$) satisfy $x_{ij} = 1$ when capabilities $i$ and $j$ are jointly optimized, and $0$ otherwise.
\end{definition}

Based on the above definition, each model $\mathcal{M}_k$ associates a binary \emph{capability indicator vector}
\begin{equation}
\mathbf{x}_{\mathcal{M}_k} = [x_1, x_2, x_3, x_{12}, x_{13}, x_{23}] \in \{0,1\}^6.
\end{equation}
For example, a model jointly trained for tool-use and reasoning has $\mathbf{x}_{\mathcal{M}} = [1,1,1,1,1,1]$, indicating that all individual capabilities and their pairwise interactions are active.
By corresponding different model with a fixed binary capability vector, we can perform controlled comparisons by comparing different models on the same question.

\begin{proposition}
\label{def:cedm}
Let $s^q_{\mathcal{M}}\in(0,1)$ be the expected correctness of model $\mathcal{M}$ on question $q$. Then $s_{\mathcal{M}}^{q}$ can be represented by  
\begin{equation}
\small{
s_{\mathcal{M}}^q = \sigma(\mathbf{x}_{\mathcal{M}}^\top \boldsymbol{\lambda}^q), \quad
\boldsymbol{\lambda}^q = [\lambda_1^q,\lambda_2^q,\lambda_3^q,\lambda_{12}^q,\lambda_{13}^q,\lambda_{23}^q],
}
\label{equation:sk}
\end{equation}
where $\sigma(\cdot)$ is sigmoid function, and  $\lambda_i^q$ and $\lambda_{ij}^q$ are the main and interaction effects for question $q$.
\end{proposition}

The proposition indicates that the correctness $s_{\mathcal{M}}^q$ can be represented as a function of the composition $\mathbf{x}_{\mathcal{M}}^\top \boldsymbol{\lambda}^q$, so that the  interaction coefficient directly serves our diagnostic purpose: $\lambda_{ij}^q > 0$ indicates a synergy between capability $x_{i}$ and $x_{j}$, while $\lambda_{ij}^q < 0$ indicates interference between them. 
Therefore, diagnosing whether reasoning and tool-use interference reduces to examining the sign of $\lambda_{23}^q$.

Unlike predictive regression (e.g., logistic models~\citep{hosmer2013applied}) that fits shared coefficients across samples, CEA performs \emph{per-question attribution}: for each $q$, we solve an independent system to obtain $\boldsymbol{\lambda}^q$.
This is feasible because the capability indicators are discrete and finite, so $\mathbf{x}_{\mathcal{M}}^\top \boldsymbol{\lambda}^q$ can exactly represent any probability assignment over these configurations (see Appendix~\ref{sec:appendix_theorem}).
Concretely, since $\boldsymbol{\lambda}^q\in\bbR^{6}$, it can be uniquely determined from six model variants with linearly independent $\mathbf{x}_{\mathcal{M}}$ and their corresponding $s_{\mathcal{M}}^{q}$.     



\paragraph{Solving $\boldsymbol{\lambda}^q$.} As clarified, specifying the interference requires to obtain $\boldsymbol{\lambda}^{q}$. To do so, applying the logit transform to both sides of Eq.~\ref{equation:sk} yields an equivalent system:
\begin{equation}
\small
z_{\mathcal{M}_k}^q = \log \frac{s_{\mathcal{M}_k}^q}{1 - s_{\mathcal{M}_k}^q}
= \mathbf{x}_{\mathcal{M}_k}^\top \boldsymbol{\lambda}^q .
\end{equation}
Since $\lambda^q \in\bbR^{6}$, stacking the equations from 6 different model variants gives the linear system
\begin{equation}
\mathbf{z}^q = \mathbf{X}\boldsymbol{\lambda}^q ,
\end{equation}
which is enough to specify the value of $\boldsymbol{\lambda}^{q}$. Here the $k$-th row of the design matrix $\mathbf{X}$ is an indicator vector of $\mathbf{x}_{\mathcal{M}_k}$ model $\mathcal{M}$, and the $k$-th component of $s^{q}_{\mathcal{M}}$ is the corresponding expected correctness.

\subsection{Constructing Controlled Model Variants for CEA}
\label{sec:gradient_masking}
As clarified, design matrix $\mathbf{X}$ should be full-rank  (Fig.~\ref{fig:section3_main}B). 
The key challenge is to construct model variants that realize different capability configurations $\mathbf{x}_{\cM_{k}}$, while keeping data, architecture, and hyperparameters unchanged.
Our insight is that different capabilities are captured by gradients on disjoint token subsets (e.g., reasoning or tool-using) during training. Thus, by \textbf{gradient masking}(\S~\ref{sec:training_derived}), we produce 4 models with varied indicator vectors from a shared base model.
However, training-time control alone cannot yield every configuration we need (4 v.s. 6).  
To fill this gap, we introduce \textbf{hybrid inference}, an inference-time routing scheme that composes separately trained models without parameter-level interaction (\S~\ref{sec:inference_derived}), which resulting the other 2 indicator vectors we require.

\vspace{-2mm}
\subsubsection{Training-derived Models}
\label{sec:training_derived}
As shown in Fig.~\ref{fig:section3_main}D, we start from a common pretrained model and apply gradient masking to derive three specialized variants. 

\textbf{1.Base Model $\mathcal{M}_{\text{Base}}$.}
The off-the-shelf pretrained LLM, providing only base capabilities: $\mathbf{x}_{{\text{Base}}} = [1,0,0,0,0,0].$
Starting from this model, we derive three variants via gradient masking. All variants share identical training data, architecture, and hyperparameters.

Using the token type $\ell(t)$ from \S~\ref{sec:preliminaries}, we define a binary mask $\mathbf{m} = [m_1, \dots, m_T]$ that gates per-token gradient contributions in Eq.\eqref{eq:joint_gradient}:
\begin{small}
\begin{equation}
\label{eq:masked_update}
\begin{aligned}
\nabla_{\theta} J(\theta)
\approx
\mathbb{E}_{\tau \sim \pi_{\theta}}
\Biggl[
\sum_{t=1}^{T}
\nabla_{\theta} \log \pi_{\theta}(c_t \mid c_{<t})
\, \cA(\tau)m_t
\Biggr].
\end{aligned}
\end{equation}
\end{small}
where $m_t \in \{0, 1\}$ gates whether token $c_t$ contributes to the update.
We partition indices into $\mathcal{T}_{\text{reas}} = \{t \mid \ell(t) = r\}$ and $\mathcal{T}_{\text{tool}} = \{t \mid \ell(t) = a\}$, and define three masking schemes:

\textbf{2.Reasoning-specialized model $\mathcal{M}_{\text{Reas}}$.}
It retains only reasoning gradients:
{\small
\[
m_t^{(\text{Reas})} = \mathbb{I}(t \in \mathcal{T}_{\text{reas}}), \qquad
\mathbf{x}_{\text{Reas}} = [1, 0, 1, 0, 1, 0].
\]
}

\textbf{3.Tool-specialized model $\mathcal{M}_{\text{Tool}}$.}
It retains only tool-use gradients:
{\small
\[
m_t^{(\text{Tool})} = \mathbb{I}(t \in \mathcal{T}_{\text{tool}}), \qquad
\mathbf{x}_{\text{Tool}} = [1, 1, 0, 1, 0, 0].
\]
}

\textbf{4.Unified model $\mathcal{M}_{\text{Unified}}$.}
It uses the standard ARL objective:
{\small
\[
m_t^{(\text{Uni})} = 1 \ \text{for all } t, \qquad
\mathbf{x}_{\text{Unified}} = [1, 1, 1, 1, 1, 1].
\]
}

\begin{figure*}[t!]
    \centering
    \includegraphics[width=\textwidth]{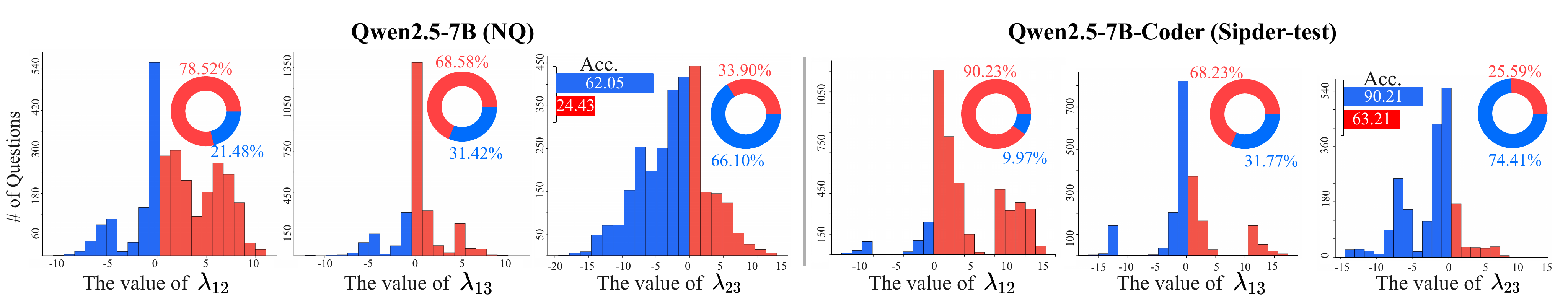}
    \vspace{-7mm}
    \caption{
    \textbf{Interference is specific to the reasoning--tool-use interaction.} 
    Blue/red indicates negative/positive interaction coefficients. $\lambda_{12}^q$ (base--reasoning) and $\lambda_{13}^q$ (base--tool) are predominantly positive, showing each capability individually synergizes with the base. In contrast, $\lambda_{23}^q$ (reasoning--tool) is dominated by negative values, revealing systematic interference under joint optimization. The \textbf{Acc.} columns show that interference concentrates on high-accuracy questions solvable by each capability alone, while synergy emerges on questions requiring both capabilities but degraded by their interference. Additional results are in Fig.~\ref{fig:other_data}.
    }
    \vspace{-2mm}   
    \label{fig:lambda23_dist}
\end{figure*}

Note that even in single-capability models (e.g., $\mathcal{M}_{\text{Reas}}$), the retained capability's gradients still update the shared base parameters, so base and that capability are jointly optimized, giving $x_{1j}=1$.

\subsubsection{Inference-derived Models}
\label{sec:inference_derived}
To complete the design matrix with 6 linearly independent rows, we add 2 hybrid-inference variants (Fig.~\ref{fig:section3_main}A) that route different token types to separately trained models, eliminating parameter-level interaction.

\textbf{5. Tool-hybrid model $\mathcal{H}_{\text{Tool}}$.}
We use the base model $\cM_{\rm{Base}}$ generates reasoning tokens; $\cM_{\rm{Tool}}$ is invoked for tool-action tokens.
Because parameters are never jointly optimized, all interaction indicators remain zero:
$\mathbf{x}_{\mathcal{H}_{\text{Tool}}} = [1,1,0,0,0,0]$.

\textbf{6. Reasoning-hybrid model $\mathcal{H}_{\text{Reas}}$.} 
Similarly, we use the reasoning  specialized model $\cM_{\rm{Reas}}$ for reasoning tokens, and the base model $\cM_{\rm{Base}}$ to handle tool-action tokens.
No joint optimization occurs, yielding
$\mathbf{x}_{\mathcal{H}_{\text{Reas}}} = [1,0,1,0,0,0]$.

By doing all of these, we obtain six rows of $\mathbf{X}\in\bbR^{6\times 6}$, and they linearly independent (Fig.~\ref{fig:section3_main}B), guaranteeing identifiability of $\boldsymbol{\lambda}^q$.

\subsection{Diagnosing Interference via CEA}
\label{sec:empirical_analysis}
We instantiate CEA to quantify all three pairwise interaction coefficients.
For each question $q$, we solve for $\blambda^q$ using the design matrix $\mathbf{X}$ induced by the six model variants defined in \S~\ref{sec:gradient_masking}.
All training-derived models are trained under identical hyper-parameters and to convergence. The correctness $s_{\mathcal{M}}^q$ of question $q$ is estimated via 50 independent answers per model--question pair.
Additional implementation details are provided in Appendix~\ref{app:interaction_details}.

Fig.~\ref{fig:lambda23_dist} shows the per-question distributions of $\lambda_{12}^q$, $\lambda_{13}^q$, and $\lambda_{23}^q$.
As expected, $\lambda_{12}^q$ (base--reasoning) and $\lambda_{13}^q$ (base--tool) are predominantly positive: adding either capability alone improves the base model.
In contrast, $\lambda_{23}^q$ (reasoning--tool) is predominantly negative, suggesting that \textbf{joint optimization of the two capabilities introduces negative interaction}.
The Acc.\ columns further reveal that interference concentrates on high-accuracy questions solvable by each capability alone, while synergy concentrates on questions that demand both capabilities yet remain low-accuracy due to their interference.
However, CEA only reveals \emph{where} and \emph{how much} interference exists, not \emph{why it arises}.
To uncover the underlying mechanism, we turn to gradient analysis.

\subsection{Explaining Interference via Gradient}
\label{sec:grad_conflict}
A natural candidate is the gradient: if reasoning and tool-use tokens require misaligned parameter updates, joint optimization cannot satisfy both simultaneously.
To test this, we measure the angular alignment between gradients from different token types.
Specifically, we compute the gradient $\mathbf{g}_{\tau}^{(b)}$ for token type $b \in \{r, a\}$ in trajectory $\tau$ across $N=16$ rollouts. We calculate the average angle between gradients of different types within the same trajectory $\mathbb{E}_{i} [\angle(\mathbf{g}_{\tau_i}^{(r)}, \mathbf{g}_{\tau_i}^{(a)})]$. 
As a baseline, we calculate the average angle between gradients of the same type from different trajectories $\mathbb{E}_{i \neq j} [\angle(\mathbf{g}_{\tau_i}^{(b)},\mathbf{g}_{\tau_j}^{(b)})]$. 
Additional implementation details are provided in Appendix~\ref{app:grad_angle}.

As shown in Fig.~\ref{fig:grad}, the angles between same-type gradients are small, while the gradients in different types (reasoning and tool-use) are nearly orthogonal.
\begin{figure}[t]
    \centering
    \includegraphics[width=\columnwidth]{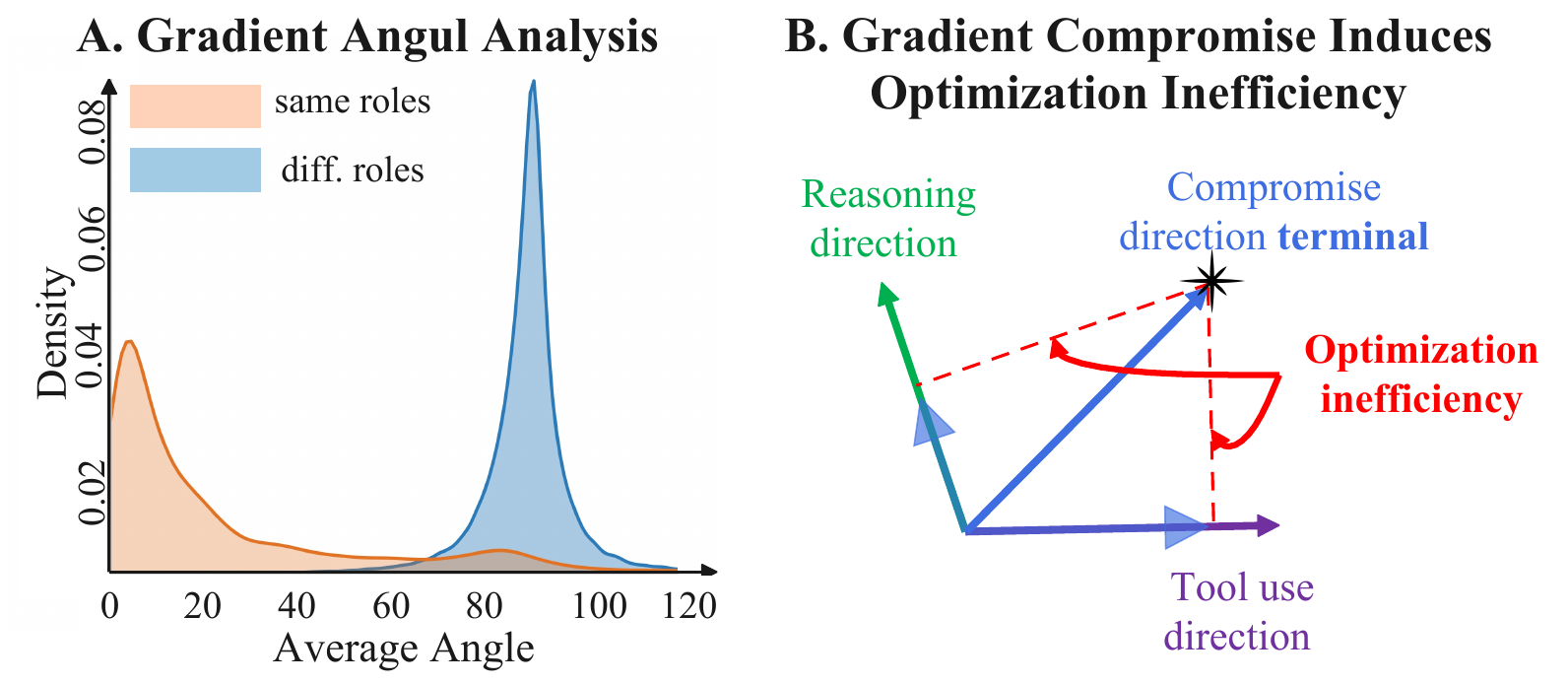}
    \caption{
    \textbf{Gradient misalignment leads to optimization inefficiency.}
\textbf{(A).} Gradient angle distributions on NQ under Qwen2.5-3B, where same-capability gradients are aligned, while reasoning and tool-use gradients are nearly orthogonal.
\textbf{(B).} Averaged orthogonal gradients yield a compromise update direction, leading to optimization inefficiency.
    }
    \vspace{-3mm}
    \label{fig:grad}
\end{figure}
This orthogonality indicates that \textit{reasoning and tool-use each require a distinct update direction}.
Consequently, averaging these gradients forces the update toward a \textbf{compromise direction} sub-optimal for both.
This finding points to a clear design principle: an effective solution must route reasoning and tool-use gradients into separate parameter subspaces.

\vspace{-2mm}
\section{Disentangled Action-Reasoning Tuning}
\label{sec:method}
Our finding (\S~\ref{sec:diagnosis}) shows that reasoning and tool-use negatively interact (Fig.~\ref{fig:lambda23_dist}) because their gradients are misaligned in shared parameter space (Fig.~\ref{fig:grad}).
This points to a clear principle: the two capabilities should update \textbf{separate parameter subspaces}.
A naive approach, training two independent models as a 2-Agent system, achieves this isolation but introduces additional system complexity (see Appendix~\ref{app:efficiency}).

To avoid this overhead while still validating our finding, we design Disentangled Action-Reasoning Tuning (DART), a \textbf{simple yet effective} framework that achieves gradient isolation within a single model.
Concretely, DART keeps the pretrained backbone weights $W$ frozen and attaches two disjoint LoRA adapters: $\theta^{r}=\{B_r, A_r\}$ for reasoning and $\theta^{a}=\{B_a, A_a\}$ for tool-use.
Freezing $W$ is necessary because if $W$ is also trained, both adapters' gradients would flow into the same backbone, bringing back the interference we aim to eliminate.

\begin{figure}[t]
    \centering
    \includegraphics[width=\columnwidth]{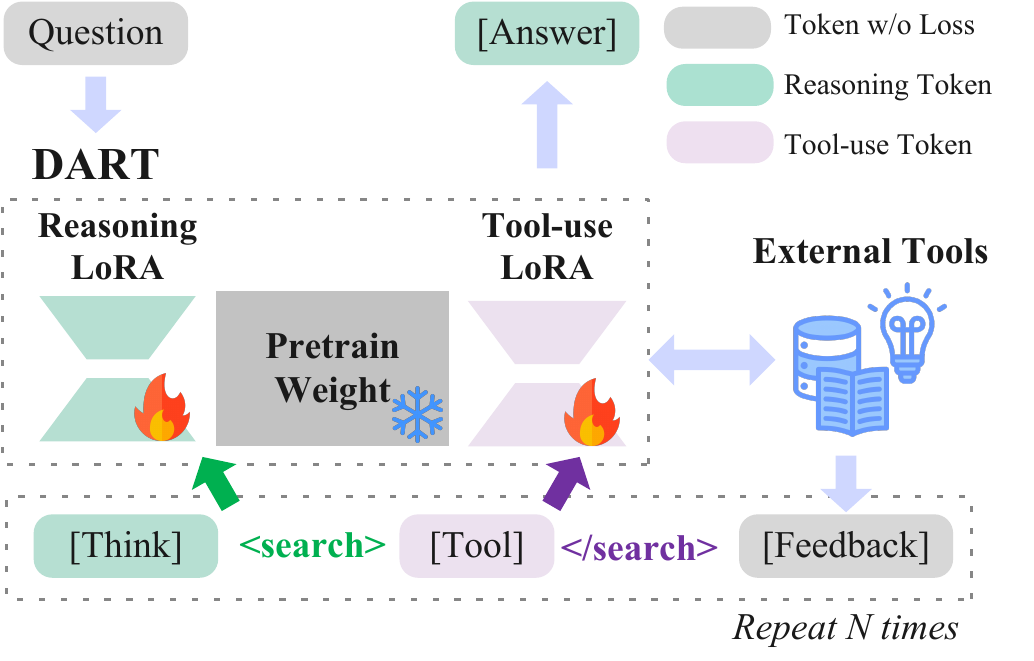}
    \caption{
    \textbf{Illustration of DART.}
    A frozen backbone augmented with two disjoint LoRA adapters for reasoning and tool-use, both attached to all linear layers, where a token-level router directs gradients into separate parameter subspaces to avoid interference.
    }
    \vspace{-4mm}
    \label{fig:dart_router}
\end{figure}

With this architecture, at each decoding step $t$, the model activates an adapter $u_t \in \{r, a\}$ determined by the token router $\ell(t)$ defined in Eq.\eqref{eq:token_router}.
As illustrated in Fig.~\ref{fig:dart_router}, token roles are assigned by special sentinel tokens (e.g., \texttt{<search>} triggers the tool-use LoRA).
To ensure robust routing, we extend the vocabulary so that each sentinel token is a single atomic entry, allowing the model to predict the routing signal as an indivisible unit rather than assembling it character by character.
Keeping the router rule-based removes additional training signals from the router itself, so that any improvement observed in subsequent experiments can be attributed purely to gradient disentanglement.

The forward pass for the hidden state $\mathbf{h}_t$ is then computed as:
\begin{equation}
\vspace{-1mm}
\mathbf{h}'_t = W \mathbf{h}_t + B_{u_t} A_{u_t} \mathbf{h}_t.
\label{eq:dart_forward}
\end{equation}
As each token activates only the adapter associated with its capability type, the two parameter sets $\theta^{r}$ and $\theta^{a}$ are updated independently.

In \S~\ref{sec:diagnosis}, we identified the root cause of the negative $\lambda_{23}^q$: reasoning and tool-use gradients compete for the same shared parameters.
DART directly addresses this cause: since $\theta^r \cap \theta^a = \emptyset$ and $W$ is frozen, no trainable parameter receives gradients from both token types. 
As we show in \S~\ref{sec:experiments}, even this minimal design already brings clear gains, confirming that gradient misalignment between reasoning and tool-use is a non-negligible limitation in ARL.

\textbf{Remark 5.1.}
One might worry that freezing $W$ limits DART's representational capacity relative to full-parameter ARL.
However, recent studies show that RL-based tuning primarily updates a sparse subset of parameters~\cite{mukherjee2025reinforcement}; 
moreover, the outcome-based reward used in ARL carries at most one bit of information per rollout, making the effective per-step learning signal extremely sparse. 
This further reduces the parameter capacity needed for each update. 
Indeed, LoRA adapters have been shown to match full-parameter RL tuning under this regime~\cite{schulman2025lora}.

\vspace{-1mm}
\section{Experiments}
\label{sec:experiments}
To verify whether the gradient misalignment between reasoning and tool-use is a non-negligible bottleneck in ARL, we evaluate our DART on seven retrieval-augmented QA benchmarks and six NL2SQL  benchmarks.
Following Search-R1~\cite{jin2025searchr} and SkyRL-SQL~\cite{liu2025skyrlsql}, all methods share the same backbone, data, RL algorithm, and hyperparameters; only gradient disentanglement differs (details in Appendix~\ref{app:exp_settings}).
We organize experiments around three questions:
\textbf{Q1}: Does resolving the identified interference by our DART yields consistent improvements across tasks and scales? (\S~\ref{sec:main_results})
\textbf{Q2}: Does interference harm both reasoning and tool-use separately, and can it be resolved by disentangling gradient as our DART did? (\S~\ref{sec:mechanism})
\textbf{Q3}: Do the gains come from gradient disentanglement or simply from increased parameter capacity? (\S~\ref{sec:ablation})

\textbf{Datasets.}
The benchmarks span two cases.
\textbf{General QA}: NQ~\cite{kwiatkowski2019natural}, TriviaQA~\cite{joshi2017triviaqa}, and PopQA~\cite{mallen2022not} for factual single-step QA.
\textbf{Multi-Hop QA}: HotpotQA~\cite{yang2018hotpotqa}, 2WikiMultiHopQA~\cite{ho2020constructing}, Musique~\cite{trivedi2022musique}, and Bamboogle~\cite{press2023measuring}, are multi-document reasoning.
We train on merged NQ and HotpotQA splits, evaluate on all seven benchmarks, and report Exact Match (EM)~\cite{yu2024rankrag}. NL2SQL datasets are in Appendix~\ref{app:eval_benchmarks}.
\vspace{-1mm}
\subsection{Main Results}\label{sec:main_results}

We first address \textbf{Q1} by comparing DART against joint-optimization and multi-LoRA baselines.

\begin{table*}[t!]
\centering
\caption{General and Multi-Hop QA results for \textbf{Qwen2.5-3b-Base/Instruct}. 
Best results are in bold.
$^{\lozenge}$ denotes results from~\citep{jin2025searchr}; $^{\dagger}$ and $^{\star}$ denote in- and out-domain datasets.}
\vspace{-2mm}
\small
\resizebox{\textwidth}{!}{
\setlength{\tabcolsep}{3pt}
\begin{tabular}{l|ccc|c|cccc|c|c}
\toprule
\multirow{2}{*}{\textbf{Methods}} 
    & \multicolumn{3}{c|}{\textbf{General QA}} 
    &  \multirow{2}{*}{\textbf{Gen-Avg}}
    & \multicolumn{4}{c|}{\textbf{Multi-Hop QA}} 
    & \multirow{2}{*}{\textbf{MH-Avg}}
    & \multirow{2}{*}{\textbf{Avg}} \\
 & NQ$^{\dagger}$ & TriviaQA$^{\star}$ & PopQA$^{\star}$ 
 & 
 & HotpotQA$^{\dagger}$ & 2Wiki$^{\star}$ & Musique$^{\star}$ & Bamboogle$^{\star}$
 & 
 & \\
\midrule
Direct Inference$^{\lozenge}$     & 0.106 & 0.288 & 0.108 & 0.167 & 0.149 & 0.244 & 0.020 & 0.024 & 0.109 & 0.134 \\
CoT$^{\lozenge}$                  & 0.023 & 0.032 & 0.005 & 0.020 & 0.021 & 0.021 & 0.002 & 0.000 & 0.011 & 0.015 \\
IRCoT$^{\lozenge}$                & 0.111 & 0.312 & 0.200 & 0.208 & 0.164 & 0.171 & 0.067 & 0.240 & 0.161 & 0.181 \\
RAG$^{\lozenge}$                  & 0.348 & 0.544 & 0.387 & 0.426 & 0.255 & 0.226 & 0.047 & 0.080 & 0.152 & 0.270 \\
SFT$^{\lozenge}$                  & 0.249 & 0.292 & 0.104 & 0.215 & 0.186 & 0.248 & 0.044 & 0.112 & 0.147 & 0.176 \\
R1-base$^{\lozenge}$              & 0.226 & 0.455 & 0.173 & 0.285 & 0.201 & 0.268 & 0.055 & 0.224 & 0.187 & 0.229 \\
R1-instruct$^{\lozenge}$          & 0.210 & 0.449 & 0.171 & 0.277 & 0.208 & 0.275 & 0.060 & 0.192 & 0.184 & 0.224 \\
Rejection Sampling$^{\lozenge}$   & 0.294 & 0.488 & 0.332 & 0.371 & 0.240 & 0.233 & 0.059 & 0.210 & 0.186 & 0.265 \\
\midrule
\multicolumn{11}{c}{Qwen2.5-3B-Instruct} \\
\midrule
Search-R1    & 0.397 & 0.565 & 0.391 & 0.451 & 0.331 & 0.310 & 0.124 & 0.232 & 0.249 & 0.336 \\
MixLoRA & 0.431 &	0.578	& 0.419	 & 0.476	& 0.346	& 0.348	& 0.125	& 0.288	& 0.277	& 0.362 \\
DART             & \textbf{0.451}          &\textbf{ 0.602 }         & \textbf{0.476} &\textbf{0.510} & \textbf{0.392}          & \textbf{0.376}          & \textbf{0.143}         & \textbf{0.352}          & \textbf{0.316}         & \textbf{0.399}\\
\midrule
\multicolumn{11}{c}{Qwen2.5-3B-Base} \\
\midrule
Search-R1       & 0.440 & 0.582 & 0.413 & 0.478 & 0.265 & 0.244 & 0.061 & 0.113 & 0.171 & 0.303 \\
MixLoRA & 0.427 &	0.595	& 0.443	& 0.488	& 0.292	& 0.282	&0.063	& 0.176	& 0.203	& 0.325 \\
DART     & \textbf{0.457}          & \textbf{0.605}          & \textbf{0.478} & \textbf{0.513} & \textbf{0.399}          & \textbf{0.389}          & \textbf{0.155}          & \textbf{0.352}          & \textbf{0.324}         & \textbf{0.405} \\
\bottomrule
\end{tabular}
}
\label{tab:qa_results_3b}
\vspace{-3mm}
\end{table*}

\textbf{Baselines.} 
\textbf{(1) Joint-optimization ARL}: Search-R1-GRPO~\cite{jin2025searchr}, the standard agentic RL baseline where reasoning and tool-use share all trainable parameters.
\textbf{(2) Soft-routing multi-LoRA}: MixLoRA~\cite{li2024mixlora}, which uses multiple LoRA experts with learned routing, partially mixing gradients across capabilities.
\textbf{(3) Other baselines}: Direct Inference, CoT~\cite{wei2022chain}, Rejection Sampling~\cite{ahn2024large}, IRCoT~\cite{trivedi2023interleaving}, RAG~\cite{lewis2020retrieval}, SFT~\cite{chung2024scaling}, and R1 variants~\cite{guo2025deepseek}.

As shown in Tab.~\ref{tab:qa_results_3b},  disentangling gradients by our DART yields consistent gains across all benchmarks.
\textbf{(1)} DART surpasses all joint-optimization baselines, confirming that \textit{the interference identified in \S~\ref{sec:diagnosis} is indeed harmful and can be effectively mitigated}.
\textbf{(2)} MixLoRA's soft routing consistently underperforms DART, indicating that \textit{partial gradient disentanglement is insufficient}; full isolation via hard token-level routing is necessary.
The same trends hold on the 7-14B backbone, Llama3.1-8B, and NL2SQL task (Appendix.~\ref{app:additional_results}).

\vspace{-1mm}
\subsection{Mechanism Analysis}\label{sec:mechanism}

\begin{figure}[t]
    \centering
    \includegraphics[width=\columnwidth]{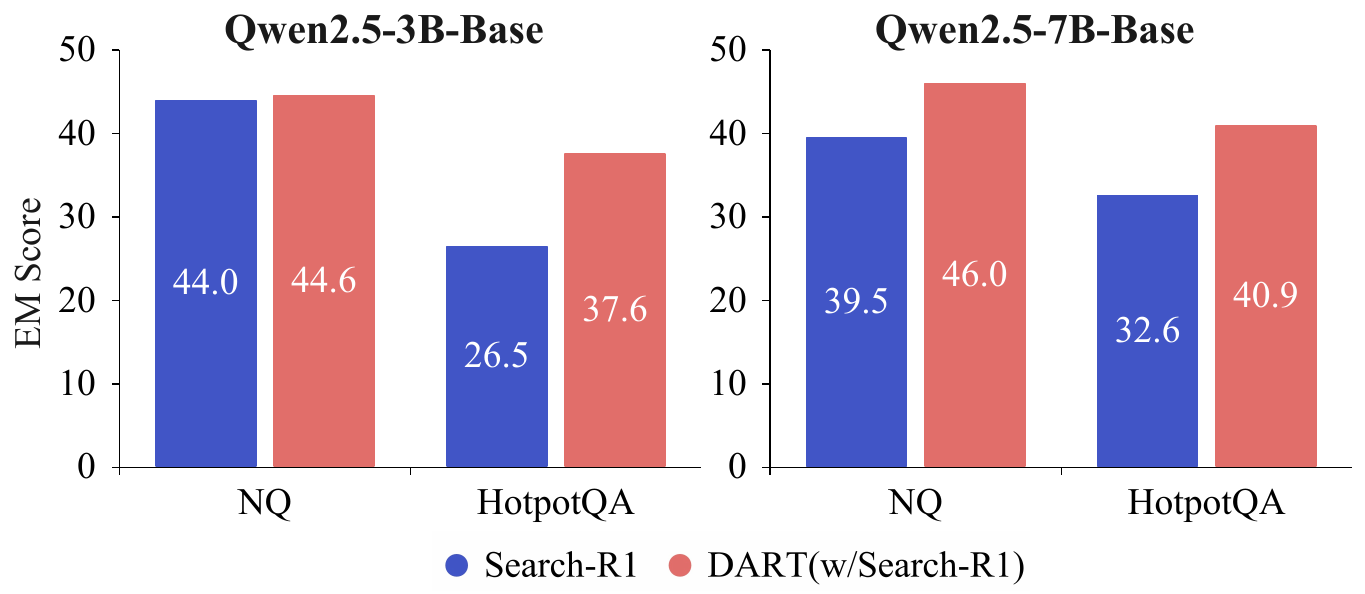}
    \caption{
    \textbf{Reasoning under Fixed Retrieval.}
    DART achieves higher EM than Search-R1 on NQ and HotpotQA when both use identical retrieval contexts, demonstrating improved reasoning  capability independent of retrieval quality.
    }
    \vspace{-6mm}
    \label{fig:fixed_context}
\end{figure}


We next explore \textbf{Q2} by two  experiments: \textbf{(1).} isolating each capability to verify that disentanglement improves both reasoning and tool-use separately; \textbf{(2).} testing whether inference-time model composition can replace training-time disentanglement.

\textbf{1. Interference Harms Each Capability Individually.}
To determine whether joint optimization degrades reasoning, we feed both DART and Search-R1 with \textbf{identical retrieval contexts} (collected from Search-R1's rollouts), so that any performance differences are attributed to reasoning ability.
As in Fig.~\ref{fig:fixed_context}, DART consistently achieves higher EM than Search-R1 given the same retrieved evidence, confirming that joint optimization impairs reasoning learning and our DART mitigates this impairment.
We further compare retrieval accuracy between DART and Search-R1 with fixed reasoning traces, confirming that joint optimization also impairs tool-use learning (Appendix~\ref{app:retrieval_accuracy}).

\textbf{2. Training-Time Isolation vs. Inference-Time Composition.}
We test whether the damage from interference can be recovered at inference time by combining separately trained specialized models.
Concretely, we extract each adapter from DART individually:  $\text{DART}_{\text{Reas}}$: reasoning adapter only, $\text{DART}_{\text{Tool}}$: tool-use adapter only, which actives the adapter only if reasoning or tool-using respectively. Then we compare them against the hybrid schemes ($\mathcal{H}_{\text{Reas}}$, $\mathcal{H}_{\text{Tool}}$) from \S~\ref{sec:gradient_masking} (Fig.~\ref{fig:section3_main}A), which compose separately trained specialized models at inference time.
As shown in Tab.~\ref{tab:dart_vs_hybrid}, each DART adapter substantially outperforms its hybrid counterpart, confirming that joint training degrades each capability and that inference-time composition \emph{cannot} recover this degradation.   

\begin{table}[h]
\centering
\caption{DART adapters vs. hybrids.}
\vspace{-2mm}
\label{tab:dart_vs_hybrid}
\small
\setlength{\tabcolsep}{6pt}
\begin{tabular}{lcccc}
\toprule
                            & \multicolumn{2}{c}{Qwen2.5-3B} & \multicolumn{2}{c}{Qwen2.5-7B} \\ 
Methods                     & NQ          & HotpotQA         & NQ          & HotpotQA         \\ \midrule
$\mathcal{H}_{\text{Reas}}$ & 0.435          & 0.324         & 0.438          & 0.327               \\
$\text{DART}_{\text{Reas}}$ & 0.448         & 0.359          & 0.449          & 0.412        \\
\rowcolor[HTML]{EFEFEF} 
$\mathcal{H}_{\text{Tool}}$ & 0.248          & 0.212         & 0.305          & 0.255              \\
\rowcolor[HTML]{EFEFEF} 
$\text{DART}_{\text{Tool}}$ & 0.372         & 0.283          & 0.378         & 0.332               \\ \bottomrule
\end{tabular}
\vspace{-4mm}
\end{table}

\vspace{-1mm} 
\subsection{Ablation Study}\label{sec:ablation}

Finally, we address \textbf{Q3} with two ablations: one compares DART with a matched-rank single LoRA and a 2-Agent upper bound to rule out capacity effects; the other sweeps LoRA rank to verify insensitivity to parameter budget.

\textbf{Ablation 1: Disentanglement vs. More Parameters.}
We verify that the gains stem from gradient isolation, not from additional parameters.
\begin{figure}[!ht]
    \centering
    \vspace{-2mm}
    \includegraphics[width=0.9\columnwidth]{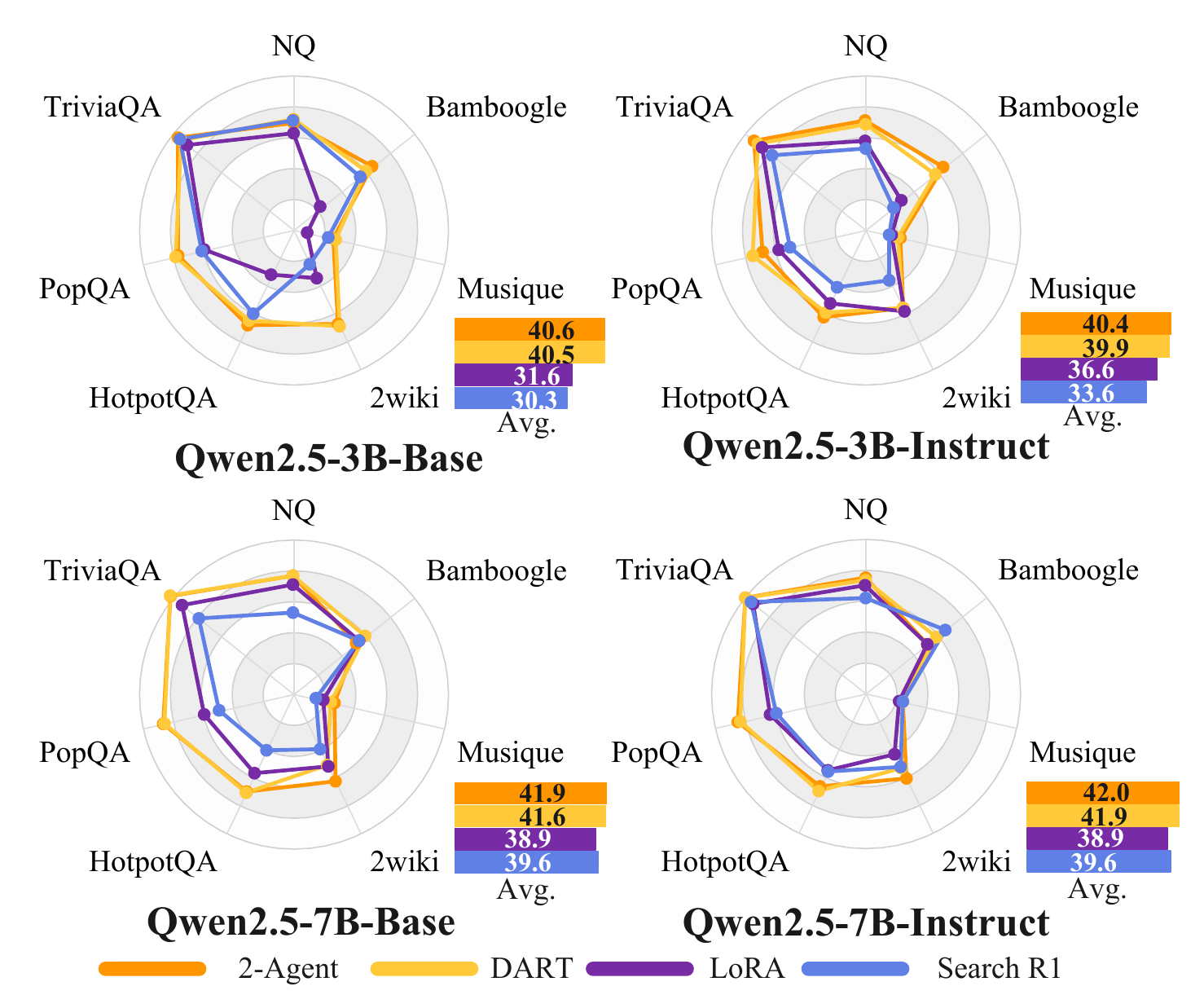}
    \caption{
    \textbf{Ablation on parameter capacity.} A matched-rank single LoRA performs similarly to Search-R1, while DART approaches the 2-Agent upper bound (two models each dedicated to one capability) across scales and benchmarks.
    }
    \label{fig:mechanism_analysis}
    \vspace{-2mm}
\end{figure}

\label{sec:mechanism_analysis}

We compare three baselines: \textbf{Search-R1} (shared parameters, full fine-tuning), \textbf{LoRA} (single adapter $r{=}16$, same total rank as DART's $r{=}8{\times}2$), and \textbf{2-Agent} (two fully independent models for tool-use and reasoning, Fig.~\ref{fig:2agent_method}), which serves as the parameter-disentangled upper bound.

As shown in Fig.~\ref{fig:mechanism_analysis}, a single LoRA with matched total rank performs nearly identically to Search-R1, indicating that extra parameters alone provide limited benefit.
In contrast, DART closely matches the 2-Agent upper bound, confirming that the gains come from isolating the two capabilities' gradients rather than from increased capacity.

\textbf{Ablation 2: Effect of LoRA Rank.}
We find that DART's performance is largely insensitive to the LoRA rank; detailed results are in Appendix~\ref{app:lora_rank}.

\section{Conclusion}
This work identifies a fundamental yet previously overlooked problem in ARL: reasoning and tool-use negatively interact under joint optimization.
CEA first exposes this phenomenon empirically, and gradient analysis traces it to misaligned parameter updates that push the two capabilities toward compromise directions.
To validate this finding, we propose DART, which simply freezes the backbone and routes reasoning and tool-use tokens to disjoint LoRA adapters, ensuring that the two capabilities should update in separate parameter subspaces.
This minimal change alone consistently surpasses all joint-optimization baselines and nearly recovers the 2-Agent upper bound across thirteen benchmarks, providing strong evidence that capability interference is a non-negligible bottleneck in current ARL systems.
As agents grow more capable, one-size-fits-all optimization may become increasingly insufficient, and capability-aware training offers a promising alternative.

\section*{Limitations}
Although resolving gradient misalignment yields consistent gains, it is not a free lunch.
Because DART only replaces the single shared LoRA with two disjoint adapters, it is fully compatible with existing RL training pipelines. However, its token-level adapter switching introduces extra scheduling complexity at high-concurrency serving.
Mitigating this overhead through batched adapter dispatch or fused multi-LoRA kernels is an engineering optimization orthogonal to DART's algorithmic contribution, which we leave to future work.


\section*{Acknowledgments}
This work was partially supported by the National Natural Science Foundation of China (Grant Nos.~62506365, 62436010, 62441230) and the Scientific Research Innovation Capability Support Project for Young Faculty (Grant No.~SRICSPYF-ZY2025001).
The authors acknowledge Beijing PARATERA Tech CO., Ltd.\ for providing HPC resources that have contributed to the research results reported within this paper. URL: \url{https://paratera.com}.


\bibliography{example_paper}

\appendix

\section{Theoretical Justification for the Capability Formulation}
\label{sec:appendix_theorem}

In this section, we demonstrate that the formulation $s_{\mathcal{M}}^q = \sigma(\mathbf{x}_{\mathcal{M}}^\top \boldsymbol{\lambda}^q)$ presented in the main text exactly characterizes the capability distribution without loss of generality.

Suppose the true success rate of question $q$ is governed by an arbitrary function $f(\cdot)$ such that $s_{\mathcal{M}}^q = \sigma(f((x_{1}, x_{2}, x_{3})))$, the $s_{\mathcal{M}}^q$ can be represented as Sigmoid function is because it is a probability in $[0, 1]$. Besides that, it is natural that the probability is only related to the abilities of base $x_{1}$, reasoning $x_{2}$, tool-using $x_{3}$.    

Because the indicator vector $(x_{1}, x_{2}, x_{3})$ consists of discrete binary variables representing specific model variants, $f(x_{1}, x_{2}, x_{3})$ only takes a \textbf{finite number} of possible values. Consequently, $f(x_{1}, x_{2}, x_{3})$ can naturally and exactly be parameterized by the inner product $\mathbf{x}_{\mathcal{M}}^\top \boldsymbol{\lambda}^q$, which is a standard result in statistics~\cite{aiken1991multiple}.

To illustrate this, suppose that $f(0, 0, 0) = 0$ without loss of generality (model without any ability can not give correct answer), then:
\begin{equation*}
\small
\begin{aligned}
& f(x_1, x_2, x_{3}) \\
&= f(1, 0, 0)x_1 + f(0, 1, 0)x_2 + f(0, 0, 1)x_{3}\\
& + (f(1, 1, 0) - f(1, 0, 0) - f(0, 1, 0))x_{1}x_{2} \\
& + (f(1, 0, 1) - f(1, 0, 0) - f(0, 0, 1))x_{1}x_{3} \\
& + (f(0, 1, 1) - f(0, 0, 1) - f(0, 1, 0))x_{2}x_{3} \\
& + (f(1, 1, 1) - f(1, 1, 0) - f(1, 0, 1) - f(0, 1, 1))x_{1}x_{2}x_{3}. 
\end{aligned}
\end{equation*}
By modeling $x_{i}x_{j}$ as $x_{ij}$ and following the observation in \citep{aiken1991multiple} that the effect of $x_{1}x_{2}x_{3}\approx 0$, we get the result that $s_{\mathcal{M}}^{q} = \sigma(f(x_{1}, x_{2}, x_{3})) = \sigma(\mathbf{x}_{\mathcal{M}}^\top \boldsymbol{\lambda}^q)$, by properly taking $\boldsymbol{\lambda}^q$, e.g., $\lambda^q_{1} = f(1, 0, 0)$.
According to the analysis in above, evaluating the diagnostic coefficients $\boldsymbol{\lambda}^q$ comprehensively captures the underlying interaction dynamics (e.g., synergy or interference) without introducing any restrictive structural assumptions.
\section{Experimental Settings}
\label{app:exp_settings}

This section details the experimental settings used in our RL training, including the training algorithm, rollout configuration, prompt templates, reward formulations, and system-level optimizations. Unless otherwise stated, these settings are shared across all experiments.

\subsection{RL Training Setup}
\label{app:rl_setup}

For GRPO training, we follow the implementation in Verl~\cite{sheng2025hybridflow}.
The backbone model is Qwen2.5~\cite{team2024qwen2} series.
For retrieval-augmented QA, we integrate an E5 retriever~\cite{wang2022text} and the 2018 Wikipedia dump~\cite{karpukhin2020dense} as the corpus.
For Natural language-to-SQL (NL2SQL) task, we use the SkyRL-SQL training set~\citep{liu2025skyrlsql} with Qwen2.5-7B-Coder as the base model, and the agent invokes a SQLite execution engine as the tool.
All experiments are conducted on a cluster of $8\times$ NVIDIA A800 GPUs.
Training is conducted for 100 optimization steps with a learning-rate warm-up ratio of 0.1.
All GRPO experiments use a fixed configuration with rollout batch size 256, gradient batch size 64, temperature 1.0, top-$p$ 1.0, and learning rate $1\times10^{-6}$.
The KL-divergence coefficient $\beta$ and clipping ratio $\epsilon$ are set to 0.001 and 0.2, respectively.

For all variants involving LoRA adaptation, we scale the learning rate by $10\times$ following prior guidance~\cite{schulman2025lora}.
To enable precise token-level routing in DART, we extend the tokenizer vocabulary with a small set of special tokens that explicitly mark reasoning and tool-use segments.

To improve training efficiency, we enable gradient checkpointing, FSDP offloading, and vLLM-based rollouts.
Model checkpoints are saved every 20 training steps.
If training diverges, we evaluate the most recent stable checkpoint according to the reward curve; otherwise, the final checkpoint is used for evaluation.

For both tasks, all compared methods share exactly the same training data, base model, prompt template, tool interface, and RL hyperparameters; the \textit{only} difference is the degree and manner of capability decoupling.
This unified setup ensures that \textit{observed differences are attributable to the parameterization and routing design}, rather than changes in data, tools, or optimization settings.

\subsection{Prompt Templates}
\label{app:prompt_template}

We adopt task-specific prompt templates that enforce a minimal structural format while avoiding content-specific biases.
Importantly, we intentionally restrict the constraints to the high-level structure (reasoning, tool invocation, final output), without enforcing reflective reasoning styles or problem-solving heuristics.
This design choice ensures that the model's learning dynamics during RL remain observable and unbiased, allowing behaviors to emerge naturally from optimization rather than prompt engineering.

\paragraph{Retrieval-Augmented QA.}
Following Search-R1~\citep{jin2025searchr}, the template structures the model output into three iterative stages: (1) a reasoning phase, (2) a search engine invocation phase, and (3) a final answer.
The maximum action budget $B$ is set to 4, and the top 3 retrieved passages are used by default.
The full template is shown in Tab.~\ref{tab:searchr1_template}.

\begin{table*}[ht]
\centering
\caption{Prompt template for retrieval-augmented QA.}
\small
\begin{tabular}{p{0.9\textwidth}}
\toprule
\textbf{Retrieval-Augmented QA Prompt Template} \\
\midrule
Answer the given question. You must conduct reasoning inside \think{and} first every time you get new information.
After reasoning, if you find you lack some knowledge, you can call a search engine by \search{query}, and it will return the top searched results between \info{and}.
You can search as many times as you want.
If you find no further external knowledge needed, you can directly provide the answer inside \answer{and} without detailed illustrations.
For example, \answer{xxx}.
Question: \texttt{question}. \\
\bottomrule
\end{tabular}
\label{tab:searchr1_template}
\end{table*}

\paragraph{Multi-Turn SQL Query Generation.}
Following SkyRL-SQL~\citep{liu2025skyrlsql}, the template structures the model output into iterative stages: (1) a reasoning phase inside \texttt{<think>} blocks, (2) a SQL tool invocation phase inside \texttt{<sql>} blocks with execution feedback returned in \texttt{<observation>} blocks, and (3) a final SQL solution.
The maximum action budget $B$ is set to 4.
The full template is shown in Tab.~\ref{tab:sql_template}.

\begin{table*}[ht]
\centering
\caption{Prompt template for multi-turn SQL query generation.}
\small
\begin{tabular}{p{0.9\textwidth}}
\toprule
\textbf{NL2SQL Prompt Template} \\
\midrule
\textbf{Task Overview:} You are a data science expert. Below, you are provided with a database schema and a natural language question. Your task is to understand the schema and generate a valid SQL query to answer the question within limited turns. You should breakdown the problem, draft your reasoning process, and generate the solution. \\[4pt]
\textbf{Database Engine:} SQLite \\[4pt]
\textbf{Database Schema:} \texttt{\{db\_details\}} \\
This schema describes the database's structure, including tables, columns, primary keys, foreign keys, and any relevant relationships or constraints. \\[4pt]
\textbf{External Knowledge:} \texttt{\{external\_knowledge\}} \\[4pt]
\textbf{Question:} \texttt{\{question\}} \\[4pt]
\textbf{Instructions:} \\
- Make sure you only output the information that is asked in the question. \\
- The generated query should return all of the information asked in the question without any missing or extra information. \\
- Before generating the final SQL query, please think through the steps of how to write the query. \\[4pt]
\textbf{Format:} \\
- Conduct thinking inside \think{...} blocks every time you get new observation or information. \\
- You can use SQL tool written within a single \sql{...} block to explore or verify. SQL tool output will be shown as dataframe inside \info{...}. Based on this observation, you can think again and refine. \\
- If you find no further exploration is needed or reaches max turns, you MUST directly provide the final SQL query solution inside \solution{...}. \\
\bottomrule
\end{tabular}
\label{tab:sql_template}
\end{table*}

\subsection{Reward Functions}
\label{app:reward}

The reward function serves as the sole training signal in our RL framework.
We adopt rule-based outcome rewards that evaluate the correctness of the model's final output, without incorporating intermediate or format-based rewards.

\paragraph{Retrieval-Augmented QA.}
For factual reasoning tasks, the reward is computed using exact match (EM):
\begin{equation}
r_{\text{qa}}(x,y) = \mathrm{EM}(a_{\text{pred}}, a_{\text{gold}}),
\end{equation}
where $a_{\text{pred}}$ is the extracted final answer from the model response $y$, and $a_{\text{gold}}$ denotes the ground-truth answer.

\paragraph{Multi-Turn SQL Query Generation.}
For NL2SQL, the reward focuses solely on execution accuracy:
\begin{equation}
R_{\text{sql}}(\mathbf{x}, \mathbf{y}) =
\begin{cases}
1  & \text{if } \mathrm{match}(\mathbf{y}, \mathbf{y}_g) \\
-1 & \text{otherwise}
\end{cases}
\end{equation}
where $\mathbf{y}$ denotes the execution result of the predicted SQL query and $\mathbf{y}_g$ denotes the ground-truth execution result.
A match is determined by comparing the execution outputs (i.e., result sets) rather than the SQL strings themselves.

\subsection{Evaluation Benchmarks for NL2SQL}
\label{app:eval_benchmarks}

Our training set is identical to that of SkyRL-SQL~\citep{liu2025skyrlsql}.
Following standard conventions, we evaluate execution accuracy (EX) on BIRD-Dev~\citep{li2023can}, \textsc{Spider}-1.0~\citep{yu2018spider}, \textsc{Spider}-DK~\citep{gan2021exploring}, \textsc{Spider}-Realistic~\citep{deng2021structure}, and \textsc{Spider}-Syn~\citep{gan2021towards}.

\section{Additional Main Results}
\label{app:additional_results}

\paragraph{Retrieval-Augmented QA with Qwen2.5-7B and Llama3.1-8B.}
Tab.~\ref{tab:qa_results_7b} reports the full results on the Qwen2.5-7B and Llama3.1-8B backbone.
The trends are consistent with those observed for the 3B model in Tab.~\ref{tab:qa_results_3b}: DART outperforms Search-R1 and MixLoRA on nearly all benchmarks and all aggregate metrics.
These results confirm that reasoning--tool-use interference is a general phenomenon independent of model scale and architecture, and that DART's disentanglement remains effective across different model families.

\begin{table*}[ht]
\centering
\small
\caption{General and Multi-Hop QA results for \textbf{Qwen2.5-7b-Base/Instruct} and \textbf{Llama3.1-8B}. 
Best results are in bold.
$^{\lozenge}$ denotes results from~\citep{jin2025searchr}; $^{\dagger}$ and $^{\star}$ denote in- and out-domain datasets.}
\vspace{-2mm}
\resizebox{\textwidth}{!}{
\setlength{\tabcolsep}{3pt}
\begin{tabular}{l|ccc|c|cccc|c|c}
\toprule
\multirow{2}{*}{\textbf{Methods}} 
    & \multicolumn{3}{c|}{\textbf{General QA}} 
    &  \multirow{2}{*}{\textbf{Gen-Avg}}
    & \multicolumn{4}{c|}{\textbf{Multi-Hop QA}} 
    &  \multirow{2}{*}{\textbf{MH-Avg}}
    & \multirow{2}{*}{\textbf{Avg}} \\
 & NQ$^{\dagger}$ & TriviaQA$^{\star}$ & PopQA$^{\star}$ 
 & 
 & HotpotQA$^{\dagger}$ & 2Wiki$^{\star}$ & Musique$^{\star}$ & Bamboogle$^{\star}$
 & 
 & \\
\midrule
Direct Inference$^{\lozenge}$     & 0.134 & 0.408 & 0.140 & 0.227 & 0.183 & 0.250 & 0.031 & 0.120 & 0.146 & 0.181 \\
CoT$^{\lozenge}$                  & 0.048 & 0.185 & 0.054 & 0.096 & 0.092 & 0.111 & 0.022 & 0.232 & 0.114 & 0.106 \\
IRCoT$^{\lozenge}$                & 0.224 & 0.478 & 0.301 & 0.334 & 0.133 & 0.149 & 0.072 & 0.224 & 0.145 & 0.239 \\
RAG$^{\lozenge}$                  & 0.349 & 0.585 & 0.392 & 0.442 & 0.299 & 0.235 & 0.058 & 0.208 & 0.200 & 0.304 \\
SFT$^{\lozenge}$                  & 0.318 & 0.354 & 0.121 & 0.264 & 0.217 & 0.259 & 0.066 & 0.112 & 0.164 & 0.207 \\
R1-base$^{\lozenge}$              & 0.297 & 0.539 & 0.199 & 0.345 & 0.242 & 0.273 & 0.083 & 0.203 & 0.200 & 0.262 \\
R1-instruct$^{\lozenge}$          & 0.270 & 0.537 & 0.199 & 0.335 & 0.237 & 0.292 & 0.072 & 0.293 & 0.224 & 0.271 \\
Rejection Sampling$^{\lozenge}$   & 0.360 & 0.592 & 0.380 & 0.444 & 0.331 & 0.296 & 0.123 & 0.355 & 0.276 & 0.348 \\
\midrule
\multicolumn{11}{c}{Qwen2.5-7B-Instruct} \\
\midrule
Search-R1 & 0.429 & 0.623 & 0.427 & 0.493 & 0.386 & 0.346 & 0.162 & \textbf{0.400} & 0.324 & 0.396 \\
MixLoRA & 0.446	& 0.625	& 0.432	& 0.501	& 0.398	 & 0.342	& 0.152	& 0.368	& 0.315	& 0.395 \\
DART      &\textbf{0.467}      & \textbf{0.642}    &\textbf{0.505} & \textbf{0.538}  & \textbf{0.431}       & \textbf{0.349} & \textbf{0.163}   & 0.397     & \textbf{0.330}  & \textbf{0.420} \\

\midrule
\multicolumn{11}{c}{Qwen2.5-7B-Base} \\
\midrule
Search-R1 & 0.395 & 0.560 & 0.388 & 0.448 & 0.326 & 0.297 & 0.125 & 0.360 & 0.277 & 0.350 \\
MixLoRA & 0.458 &	0.626	&0.443	&0.509	& 0.408	& 0.318	& 0.156	& 0.368	& 0.323	& 0.402 \\
DART   & \textbf{0.472}      & \textbf{0.639}    & \textbf{0.507} & \textbf{0.539}   & \textbf{0.425}       & \textbf{0.338} & \textbf{0.155}   & \textbf{0.376}    & \textbf{0.323} & \textbf{0.416}   \\
\midrule
\multicolumn{11}{c}{Llama3.1-8B-Instruct} \\
\midrule
Search-R1 & 0.481 & 0.659 & 0.489 & 0.543 & 0.438 & 0.387 & 0.201 & 0.448 & 0.368 & 0.443 \\
MixLoRA & 0.477 & 0.659 & 0.475 & 0.537 & 0.424 & 0.411 & 0.192 & 0.472 & 0.375 & 0.444 \\
DART & \textbf{0.501} & \textbf{0.665} & \textbf{0.516} & \textbf{0.561} & \textbf{0.464} & \textbf{0.411} & \textbf{0.217} & \textbf{0.476} & \textbf{0.392} & \textbf{0.464} \\
\bottomrule
\end{tabular}
}
\label{tab:qa_results_7b}
\vspace{-3mm}
\end{table*}

\paragraph{Multi-Turn SQL Query Generation.}
\label{app:vt_sql_results}
We further evaluate DART on the NL2SQL task to test whether its disentanglement benefit transfers to a different tool-use scenario.
The experimental setup and evaluation protocol are detailed in Appendix~\ref{app:exp_settings}.
As shown in Tab.~\ref{tab:vt_sql_results}, DART achieves the best single-model performance, surpassing the Sky-SQL baseline across all six benchmarks while closely matching the more resource-intensive 2-Agent system.
MixLoRA's soft routing again underperforms, reinforcing the finding that partial gradient disentanglement is insufficient for effective capability isolation.
We further note that MixLoRA's degradation is more severe on NL2SQL than on QA. We attribute this to the larger modality gap between reasoning tokens (natural language) and tool-use tokens (formal SQL): soft routing forces every adapter to absorb gradients from both modalities, amplifying interference when the two token distributions diverge significantly.
These results confirm that DART's disentanglement benefit generalizes from retrieval-augmented QA to the NL2SQL setting, where the tool interface, action space, and reward signal are fundamentally different.

\begin{table*}[t]
\centering
\caption{
\textbf{Execution accuracy (EX) on NL2SQL benchmarks.}
All methods use the same SkyRL-SQL training data and Qwen2.5-7B-Coder backbone; only the decoupling strategy differs.
}
\label{tab:vt_sql_results}
\small
\setlength{\tabcolsep}{6pt}
\begin{tabular}{lccccccc}
\toprule
\textbf{Method} & \textbf{BIRD-Dev} & \textbf{Spider-Dev} & \textbf{Spider-DK} & \textbf{Spider-Realistic} & \textbf{Spider-Syn} & \textbf{Spider-Test} & \textbf{Avg.} \\
\midrule
Sky-SQL     & 0.4912 & 0.8085 & 0.7121 & 0.7611 & 0.6925 & 0.8281 & 0.7156 \\
LoRA        & 0.4844 & 0.8075 & 0.7159 & 0.7598 & 0.7021 & 0.8277 & 0.7162 \\
MixLoRA     & 0.4735 & 0.7821 & 0.6820 & 0.7521 & 0.6822 & 0.7823 & 0.6924 \\
\midrule
2-Agent     & \textbf{0.5271} & 0.8162 & 0.7084 & \textbf{0.7894} & 0.7108 & \textbf{0.8314} & 0.7306 \\
DART        & 0.5215 & \textbf{0.8251} & \textbf{0.7144} & 0.7815 & \textbf{0.7137} & 0.8291 & \textbf{0.7309} \\
\bottomrule
\end{tabular}
\end{table*}

\paragraph{Scaling to a Larger and Newer Backbone (Qwen3-14B).}
To verify that both the reasoning--tool interference and DART's remedy generalize beyond the 3B--8B range and beyond the Qwen2.5 family, we additionally evaluate on \textbf{Qwen3-14B-Base}, which is at once larger in scale and drawn from a newer, distinct model family.
All three methods are run under the same protocol as Appendix~\ref{app:exp_settings}, with numbers reported as mean $\pm$ std over 3 seeds.
As shown in Tab.~\ref{tab:qwen3_14b}, the same pattern observed at 3B/7B/8B continues to hold: all methods scale up clearly over their 7B counterparts, the joint baselines Search-R1 and MixLoRA remain close (soft routing still fails to resolve the interference), while \textbf{DART leads on every benchmark by a clear margin}.
This confirms that the interference and DART's disentangled remedy generalize across both scale and architecture.

\begin{table}[t]
\centering
\caption{
\textbf{Exact-Match accuracy on retrieval-augmented QA with Qwen3-14B-Base} (mean $\pm$ std over 3 seeds). Best per column in \textbf{bold}.
}
\label{tab:qwen3_14b}
\small
\setlength{\tabcolsep}{4pt}
\resizebox{\columnwidth}{!}{%
\begin{tabular}{lccccc}
\toprule
\textbf{Method} & \textbf{NQ} & \textbf{HotpotQA} & \textbf{2Wiki} & \textbf{PopQA} & \textbf{Avg.} \\
\midrule
Search-R1 & $0.499${\scriptsize$\pm.006$} & $0.463${\scriptsize$\pm.007$} & $0.453${\scriptsize$\pm.012$} & $0.504${\scriptsize$\pm.009$} & 0.480 \\
MixLoRA   & $0.489${\scriptsize$\pm.011$} & $0.450${\scriptsize$\pm.006$} & $0.443${\scriptsize$\pm.008$} & $0.505${\scriptsize$\pm.014$} & 0.472 \\
\textbf{DART} & $\mathbf{0.515}${\scriptsize$\pm.015$} & $\mathbf{0.479}${\scriptsize$\pm.011$} & $\mathbf{0.471}${\scriptsize$\pm.005$} & $\mathbf{0.523}${\scriptsize$\pm.008$} & \textbf{0.497} \\
\bottomrule
\end{tabular}%
}
\end{table}

\section{Experimental Details and Extended Results for CEA}
\label{app:interaction_details}

This section provides additional experimental details for the CEA presented in \S~\ref{sec:empirical_analysis} and extends the analysis to both tasks to verify the generality of our findings
\subsection{Shared Experimental Protocol}
\label{app:cea_shared}

For each question $q$, we consider the six models defined in \S~\ref{sec:capability_decomposition}, which correspond to different combinations of base, tool-use, and reasoning capabilities and induce a fixed design matrix $\mathbf{X}$.
All models are trained under identical hyper-parameters and to convergence, \textit{differing only in capability activation}, which ensures controlled and fair comparisons across models.
Given the six empirical correctness estimates $\{\hat{s}_{\mathcal{M}}^q\}$, we solve a linear system in logit space to obtain the question-level effect vector $\boldsymbol{\lambda}^q$.
The interaction coefficient $\lambda_{23}^q$ captures the deviation of the jointly optimized reasoning--tool-use configuration from the additive expectation of the two capabilities, where negative values indicate interference and positive values synergy.
Similarly, $\lambda_{12}^q$ and $\lambda_{13}^q$ capture the base--reasoning and base--tool interactions, respectively.

For numerical stability, we discard any question $q$ for which all six model variants yield $\hat{s}_{\mathcal{M}_k}^q = 0$, i.e., no model ever produces a correct answer across all $N$ samples.
Including such questions would cause the smoothed logit values to cluster near $-\infty$ for every variant, making the resulting $\boldsymbol{\lambda}^q$ dominated by boundary artifacts rather than genuine capability differences.
After filtering, the additive smoothing defined in the main text ensures that all remaining $\bar{s}_{\mathcal{M}_k}^q$ lie strictly in $(0,1)$ and the logit transform is well-behaved.
We then aggregate all three pairwise coefficients across retained questions and report the proportion of negative and positive values.
In all experiments, we set $N=50$ and adopt the same stochastic decoding strategy as Appendix~\ref{app:exp_settings}, with fixed temperature and top-$p$ sampling.
Averaging over multiple samples reduces decoding noise and yields a more stable estimate of model correctness.
Except for the task-specific details described below, all inference and evaluation hyper-parameters follow the settings of the main experiments.

\subsection{Retrieval-Augmented QA}
\label{app:cea_qa}

\paragraph{Token Role Definition.}
In the retrieval-augmented QA trajectory, we define two token roles:
(1) \emph{reasoning tokens} correspond to content within \texttt{<think>} blocks, where the model analyzes retrieved information and performs multi-step reasoning;
(2) \emph{tool-use tokens} correspond to content within \texttt{<search>} blocks, where the model formulates search queries to invoke the retrieval engine.
These two roles are mutually exclusive and jointly cover all non-prompt tokens in each trajectory, enabling the six-variant design matrix $\mathbf{X}$ and gradient masking procedure described in \S~\ref{sec:gradient_masking}.

\paragraph{Correctness Metric.}
We adopt exact match (EM) as the correctness measure:
\[
\hat{s}_{\mathcal{M}}^q
=
\frac{1}{N}\sum_{n=1}^{N}\mathrm{EM}\big(a_q^{(n)}, a_{\text{gold}}\big),
\]
where $a_q^{(n)}$ is the extracted final answer from the $n$-th rollout, $a_{\text{gold}}$ is the ground-truth answer, and $\mathrm{EM}(\cdot)\in\{0,1\}$.

\subsection{Multi-Turn SQL Query Generation (NL2SQL)}
\label{app:cea_sql}

In this setting, the tool is a SQLite execution engine rather than a search engine, and the action space consists of SQL queries rather than search queries.
Following the training setup in Appendix~\ref{app:exp_settings}, we use the SkyRL-SQL training set~\citep{liu2025skyrlsql} with Qwen2.5-7B-Coder as the base model.

\paragraph{Token Role Definition.}
In the NL2SQL trajectory, we define two token roles analogous to the QA setting:
(1) \emph{reasoning tokens} correspond to content within \texttt{<think>} blocks, where the model plans query strategies, interprets execution feedback, and reasons about schema relationships;
(2) \emph{tool-use tokens} correspond to content within \texttt{<sql>} blocks, where the model generates SQL queries for execution.
These two roles are mutually exclusive and jointly cover all non-prompt tokens in each trajectory, enabling the same six-variant design matrix $\mathbf{X}$ and gradient masking procedure.

\paragraph{Correctness Metric.}
Instead of exact match, we adopt execution accuracy (EX) as the correctness measure:
\[
\hat{s}_{\mathcal{M}}^q = \frac{1}{N}\sum_{n=1}^{N} \mathrm{match}\!\left(\mathrm{exec}(y_q^{(n)}),\; \mathrm{exec}(y_{\text{gold}})\right),
\]
where $\mathrm{exec}(\cdot)$ denotes the result set returned by executing the SQL query, and $\mathrm{match}(\cdot)\in\{0,1\}$.
A prediction is correct if its execution output matches the ground-truth execution output, regardless of syntactic SQL differences.

\subsection{Extended Results}
\label{app:cea_findings}

\begin{figure*}[ht]
    \centering
    \includegraphics[width=\textwidth]{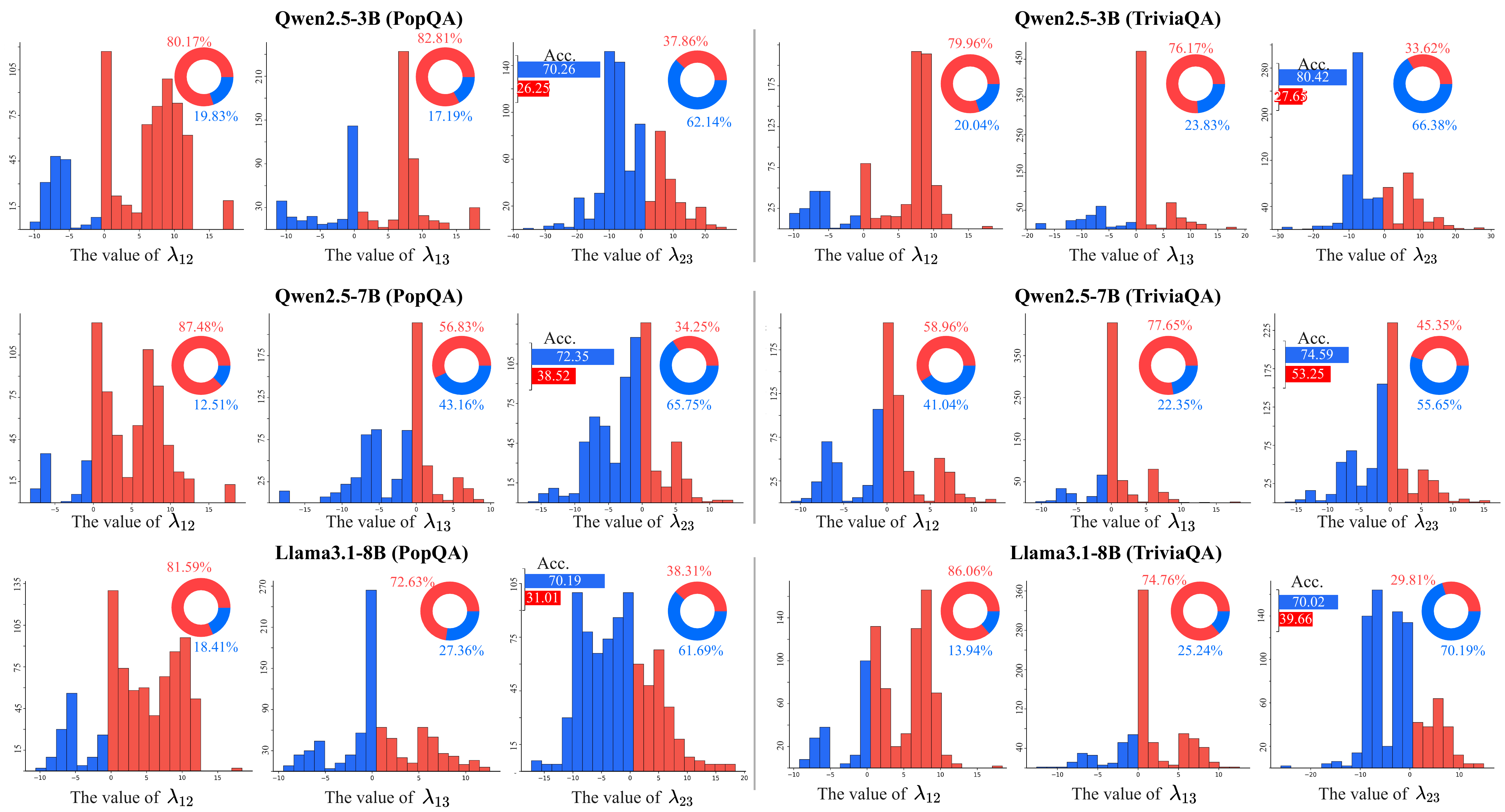}
    \caption{
    \textbf{The reasoning--tool-use interference generalizes across datasets, model scales, and architectures.}
    We extend the CEA analysis of Fig.~\ref{fig:lambda23_dist} to PopQA and TriviaQA with Qwen2.5-3B, Qwen2.5-7B, and Llama3.1-8B, evaluating on the first 1,000 test samples per dataset.
    Blue/red indicates negative/positive interaction coefficients.
    Across all combinations, $\lambda_{23}^q$ is consistently dominated by negative values. The \textbf{Acc.} columns confirm that interference concentrates on high-accuracy questions. The synergy group in 7B models shows higher accuracy than in 3B, reflecting the stronger base capacity.
    }
    \label{fig:other_data}
\end{figure*}

We extend the CEA analysis of Fig.~\ref{fig:lambda23_dist} to additional datasets (PopQA, TriviaQA) and model architectures (Qwen2.5-3B, Qwen2.5-7B, Llama3.1-8B). For each dataset, we evaluate on the first 1,000 samples from the test set.
As shown in Fig.~\ref{fig:other_data}, $\lambda_{23}^q$ is consistently dominated by negative values across all dataset--model combinations.
The Acc.\ columns confirm that interference concentrates on high-accuracy questions where both capabilities are well-learned.
Additionally, the synergy group (positive $\lambda_{23}^q$) in 7B models achieves higher accuracy than that in 3B models, reflecting the stronger base capacity.
This cross-task and cross-architecture consistency provides evidence that the interference identified by CEA reflects a general property of ARL, rather than an artifact of a specific model or dataset.

\section{Implementation Details of Gradient Misalignment}
\label{app:grad_angle}

This section provides implementation details for the gradient angle analysis described in the main text. Following the training and sampling protocol of~\cite{jin2025searchr}, for each input query we sample \(N=16\) rollouts \(\{\tau_i\}_{i=1}^{N}\) from the current policy. All analyses are conducted with fixed base model parameters: we perform forward and backward passes solely to extract gradients and do not update the model.

Based on the token-level masked update in Eq.~\ref{eq:masked_update} and the hyperparameter settings described in Appendix~\ref{app:exp_settings}, we compute policy gradients for different token roles within each trajectory. Specifically, for each rollout \(\tau_i\), we compute gradients \(\mathbf{g}_{\tau_i}^{(b)}\) for token role \(b \in \{r, a\}\), where \(r\) denotes reasoning tokens and \(a\) denotes tool-use tokens. 
Gradients for different roles are obtained via separate backward passes, with gradients explicitly zeroed between passes to avoid accumulation effects.

Gradient angles are computed from the cosine similarity between two gradient vectors. 
Given two gradients $\mathbf{g}_1$ and $\mathbf{g}_2$, we first compute their cosine similarity as
\[
\cos(\mathbf{g}_1, \mathbf{g}_2) = 
\frac{\mathbf{g}_1^\top \mathbf{g}_2}
{\|\mathbf{g}_1\|_2 \, \|\mathbf{g}_2\|_2},
\]
where gradients are flattened over all model parameters. 
The corresponding angle is then obtained by
\[
\angle(\mathbf{g}_1, \mathbf{g}_2) = \arccos\!\left(\cos(\mathbf{g}_1, \mathbf{g}_2)\right),
\]
which yields values in $[0, \pi]$. 
This conversion allows us to interpret gradient alignment geometrically, with smaller angles indicating stronger alignment and angles approaching $\pi/2$ or larger indicating increasing degrees of misalignment.

All experiments use the same numerical and system settings as training. We enable FlashAttention-2 and gradient checkpointing to support long-sequence computation, and perform all forward and backward passes in bfloat16 precision. In memory-constrained environments, parameters are managed with CPU offloading.
The maximum lengths of both prompts and responses are set to 4096 tokens. 

Notably, all gradients are computed over the full sequence, but only tokens selected by the corresponding role mask contribute to the policy loss and backpropagation. Gradient clipping is disabled by default to avoid altering the geometry of gradients. 
We additionally observe qualitatively similar gradient angle patterns when repeating the analysis at other training steps, suggesting that the observed gradient interference is not specific to a single checkpoint.

\begin{figure*}[ht]
    \centering
    \includegraphics[width=\textwidth]{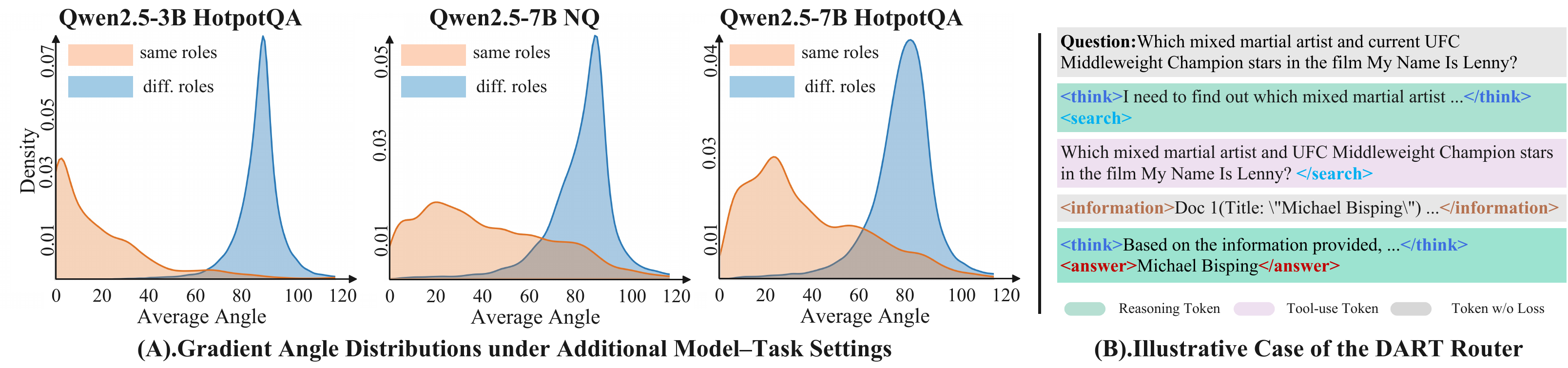}
    \caption{
    \textbf{Gradient Misalignment and Router Behavior in DART.}
\textbf{(A).} Gradient angle distributions under additional model–task settings, showing that gradients from the same capability are well aligned, while gradients between reasoning and tool-use tokens are largely orthogonal. \textbf{(B).} An illustrative example of the DART router, highlighting \textbf{rule-based} token-level routing decisions that distinguish reasoning, tool-use, and loss-free tokens during a tool-augmented QA process.
    }
    \label{fig:grad+router}
\end{figure*}

Fig.~\ref{fig:grad+router}(A) shows that across all settings, reasoning–tool gradients are close to orthogonal, while same-capability gradients exhibit stronger alignment, indicating clear directional separation.
Compared to the 3B model, the 7B model shows a more dispersed distribution of same-role gradients, which we attribute to its larger capacity: with more parameters, the model admits a wider range of gradient directions for the same capability across different samples.

\section{Theoretical Efficiency: DART vs. 2-Agent System}
\label{app:efficiency}

A common alternative to a unified model is a disentangled 2-agent system, where a specialized reasoning model $\mathcal{M}_{\text{Reas}}$ and a tool-use model $\mathcal{M}_{\text{Tool}}$ collaborate. 
While this modularity seems intuitive, it introduces significant overhead in resource consumption and latency. 
Below, we provide a theoretical analysis of why the DART framework is more efficient.

\begin{figure}[ht]
    \centering
    \includegraphics[width=\columnwidth]{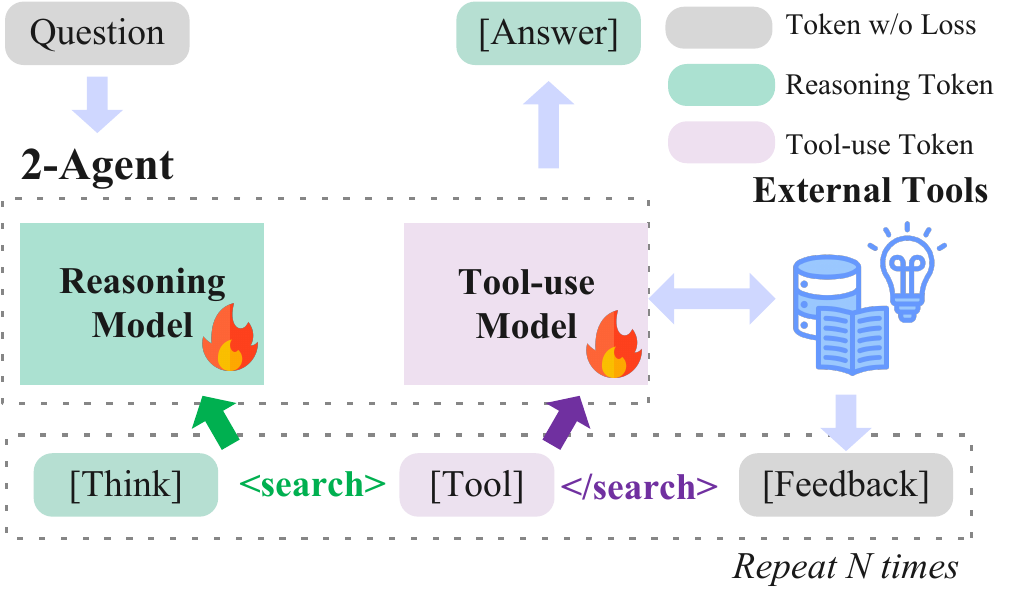}
    \caption{
    \textbf{2-Agent System Architecture.}
    A reasoning model and a tool-use model operate as separate models and interact through explicit handoffs. The reasoning model decides when to invoke tools, while the tool-use model executes tool calls and returns feedback.
    }
    \label{fig:2agent_method}
\end{figure}

\paragraph{Training Memory: The Shared-Backbone Advantage}
We analyze the training-time GPU memory complexity of DART in comparison with 2-agent system. 
Let $P$ denote the number of parameters in the backbone model, and let $p$ denote the number of parameters introduced by a LoRA adapter, where $p \ll P$ (typically below $0.5\%$ of $P$). 
Model parameters and gradients are stored in BF16 precision, while optimizer states are stored in FP32 precision.

Under a disentangled multi-agent GRPO setup, two trainable policy backbone models must be resident on GPU. For each model, training stores parameters, gradients, and Adam-style optimizer states, contributing approximately parameters. 
As a result, the dominant static memory cost scales as $\mathcal{O}(P_{\text{2-agent}}) \approx 2 \times 4P = 8P$.
In contrast, DART trains both capabilities within a single shared backbone and confines all trainable parameters to lightweight LoRA adapters.
The backbone is frozen, and gradients as well as optimizer states are stored only for the adapter parameters. 
As a result, the dominant static memory cost scales as $\mathcal{O}(P_{\text{DART}}) \approx P + \mathcal{O}(p)$, where the contribution of $p$ is negligible.

According to our empirical observation, the resulting memory ratio can be approximated as
\[
\frac{\mathcal{O}(P_{\text{2-agent}})}{\mathcal{O}(P_{\text{DART}})} \approx \mathcal{O}(8).
\]
DART reduces the training-time static memory footprint by roughly 
$8\times$ while maintaining performance comparable to 2-agent.

\begin{figure*}[t!]
    \centering
    \includegraphics[width=0.75\textwidth]{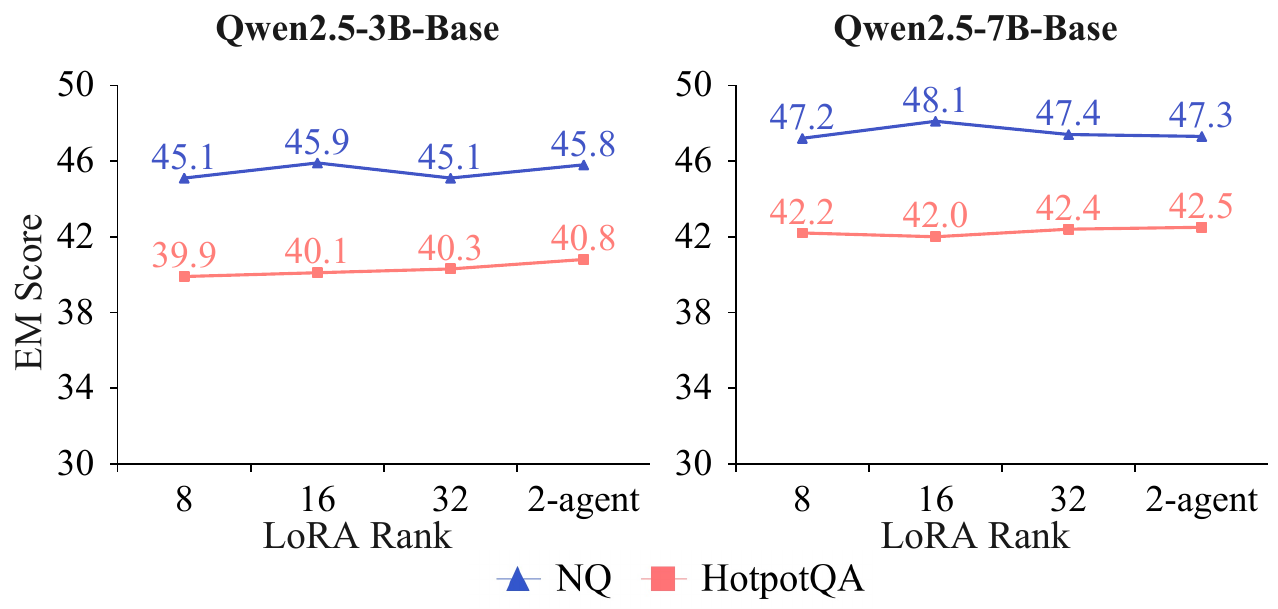}
    \caption{
    \textbf{LoRA Rank Sensitivity.}
    DART exhibits stable EM performance across LoRA ranks and remains close to the 2-agent baseline.
    }
    \label{fig:lora_rank}
\end{figure*}

\begin{table*}[t!]
\centering
\small
\begin{tabular}{lcc}
\toprule
\textbf{Metric} & \textbf{disentangled 2-Agent (LoRA)} & \textbf{DART (Multi-LoRA)} \\
\midrule
Backbone Instances & 2 & \textbf{1} \\
VRAM (Weight-dominant) & $\approx 2P$ & $\mathbf{\approx 1P}$ \\
Context Switching Cost & High (Re-encoding $\mathcal{O}(L^2)$) & \textbf{Zero (KV-Cache Reuse)} \\
\bottomrule
\end{tabular}
\caption{Theoretical comparison between a disentangled 2-agent system and the DART framework. $P$ denotes the backbone parameter count; $L$ denotes sequence length.}
\label{tab:efficiency_comp}
\end{table*}

\paragraph{Inference Latency: The KV-Cache Advantage.}
The most critical bottleneck in multi-turn interactions is computing the prefill during context switching.
\begin{itemize}
    \item \textbf{2-Agent Latency:} When $\mathcal{M}_{\text{Reas}}$ generates a thought and hands it to $\mathcal{M}_{\text{Tool}}$, the latter must re-encode the entire conversation history $H$ of length $L$ to build its own Key-Value (KV) cache. This re-computation has a complexity of $\mathcal{O}(L^2)$.
    \item \textbf{DART Latency:} Since DART operates on a single backbone, the KV-cache remains valid across capability switches. Moving from reasoning to tool-invocation only requires a negligible $\mathcal{O}(1)$ switch of the active LoRA ranks. The historical context is never re-processed, drastically reducing the Time-To-First-Token (TTFT) for subsequent turns.
\end{itemize}

As summarized in Tab.~\ref{tab:efficiency_comp}, DART simplifies the deployment stack. A 2-agent system requires an external orchestrator to synchronize states and format prompts between models, whereas DART internalizes this logic within a single inference pipeline.

\section{Effect of LoRA Rank in DART}
\label{app:lora_rank}

We study the effect of the LoRA rank in DART by varying the adapter rank on the \textbf{Qwen2.5-3B-Base} model.
Figure.~\ref{fig:lora_rank}(A) reports DART’s EM performance on NQ and HotpotQA under different LoRA ranks (8/16/32) for both Qwen2.5-3B and Qwen2.5-7B backbones, with the 2-agent system shown as a reference.
Overall, \textbf{DART is not strongly sensitive to the rank choice}: varying the rank changes EM only marginally, and the relative ordering across datasets and model scales remains consistent.
Across all settings, DART stays close to the 2-agent baseline, indicating that \textit{its improvements are not driven by simply increasing adapter capacity}. This is an interesting observation, which indicates that under the disentangled learning paradigm, a slight parameter capacity is enough to make the model completes the task well in practice.  

\begin{figure}[t]
    \centering
    \includegraphics[width=\columnwidth]{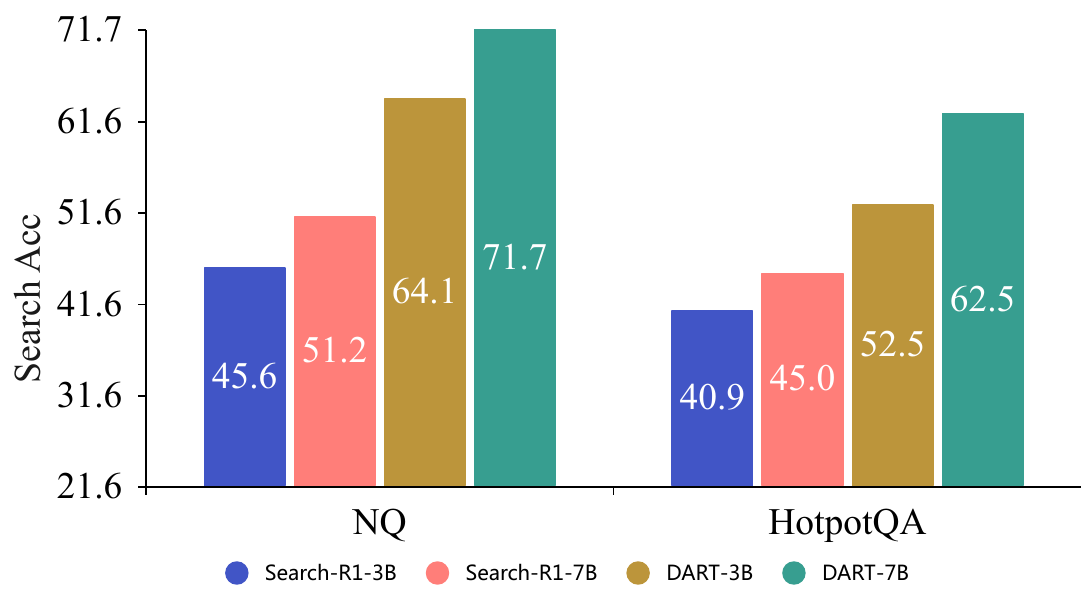}
    \caption{
    \textbf{Search Accuracy of DART.}
    DART consistently achieves higher search accuracy than Search-R1 across both datasets and model scales, and retrieval accuracy scales monotonically with model size.
    }
    \label{fig:search_acc}
\end{figure}

\section{Retrieval Accuracy Evaluation}
\label{app:retrieval_accuracy}
In section \ref{sec:mechanism}, we show that the single ability of DART is also improved, compared to the hybrid model. Next, we directly verify the search accuracy of DART model is improved, compared with baseline model. Concretely, we report the retrieval accuracy results and the corresponding evaluation protocol, which are presented exclusively here to analyze tool-use behavior under different training paradigms.
We compare the jointly trained Search-R1 baseline with DART on the NQ and HotpotQA benchmarks, focusing on the model’s ability to retrieve task-relevant information during inference.

We evaluate retrieval performance using \emph{retrieval accuracy}. 
Let $\mathcal{S}$ denote the evaluation set. 
For each example $j \in \mathcal{S}$, the model retrieves a set of information documents or passages denoted by $\mathcal{D}_j$, and the ground-truth answer set is given by $G_j$. 
We define a retrieval correctness indicator $\mathrm{RetCorrect}(\mathcal{D}_j, G_j)$, which equals $1$ if there exists at least one retrieved document in $\mathcal{D}_j$ that matches any element in $G_j$, and $0$ otherwise. 
The overall retrieval accuracy is then defined as
\[
\mathrm{Acc}
=
\frac{1}{|\mathcal{S}|}
\sum_{j \in \mathcal{S}}
\mathrm{RetCorrect}(\mathcal{D}_j, G_j).
\]

We report retrieval accuracy for both Qwen2.5-3B and Qwen2.5-7B backbones under identical data splits and inference settings.  
Search-R1 optimizes reasoning and tool use jointly, whereas DART isolates their parameter updates during training. 
All methods share the same retrieval format and correctness criterion.

As shown in Figure~\ref{fig:search_acc}, DART consistently achieves higher retrieval accuracy than Search-R1 across both datasets and model scales. 
This indicates that DART retrieves task-relevant information more reliably, particularly on multi-hop and fact-intensive tasks, highlighting the effectiveness of training-time capability disentanglement for tool use.
As expected, retrieval accuracy scales consistently with model size: the 7B backbone outperforms the 3B backbone for both Search-R1 and DART.

\end{document}